\documentclass{article}

\usepackage[numbers]{natbib}
\usepackage{microtype}
\usepackage{graphicx}
\usepackage{subcaption}
\usepackage{booktabs}
\usepackage{hyperref}
\usepackage[table,dvipsnames]{xcolor}
\usepackage{colortbl}
\usepackage{multirow}
\usepackage{pifont}
\usepackage{tcolorbox}
\tcbuselibrary{skins,breakable}

\newcommand{\cmark}{\ding{51}}
\newcommand{\xmark}{\ding{55}}

\usepackage[preprint]{icml2026}

\usepackage{algorithm}
\usepackage{algorithmic}
\usepackage{amsmath}
\usepackage{amssymb}
\usepackage{mathtools}
\usepackage{amsthm}
\usepackage{bm}

\usepackage[capitalize,noabbrev]{cleveref}

\theoremstyle{plain}

\theoremstyle{definition}

\theoremstyle{remark}

\newcommand{\method}{\textsc{DiffuSpeech}}
\newcommand{\dataset}{\textsc{ThinkingTalk}}

\newcommand{\E}{\mathbb{E}}
\newcommand{\R}{\mathbb{R}}

\definecolor{tablerow1}{RGB}{245, 248, 252}
\definecolor{tablerow2}{RGB}{255, 255, 255}
\definecolor{oursrow}{RGB}{232, 245, 233}
\definecolor{impgreen}{RGB}{34, 139, 34}
\newcommand{\gray}[1]{\textcolor{gray}{#1}}
\newcommand{\imp}[2]{\textbf{#1}{\footnotesize\textcolor{impgreen}{$\uparrow$#2}}}

\newcommand{\ours}{\rowcolor{oursrow}}

\definecolor{userbox}{RGB}{225, 237, 252}
\definecolor{thinkbox}{RGB}{255, 249, 230}
\definecolor{replybox}{RGB}{232, 245, 233}
\tcbset{
    examplebox/.style={
        enhanced,
        boxrule=0.5pt,
        colframe=black!20,
        arc=2pt,
        fonttitle=\bfseries\small,
        coltitle=black,
        attach boxed title to top left={yshift=-2mm, xshift=3mm},
        boxed title style={size=small, colback=white, boxrule=0.5pt, colframe=black!20, arc=2pt},
        left=3pt, right=3pt, top=4pt, bottom=3pt,
        breakable
    }
}

\icmltitlerunning{DiffuSpeech: Silent Thought, Spoken Answer via Unified Speech-Text Diffusion}

\begin{document}

\twocolumn[
  \icmltitle{DiffuSpeech: Silent Thought, Spoken Answer \\ via Unified Speech-Text Diffusion}

  \icmlsetsymbol{equal}{*}

  \begin{icmlauthorlist}
    \icmlauthor{Yuxuan Lou}{nus,equal}
    \icmlauthor{Ziming Wu}{tencent,equal}
    \icmlauthor{Yaochen Wang}{nus}
    \icmlauthor{Yong Liu}{nus}
    \icmlauthor{Yingxuan Ren}{nus}
    \icmlauthor{Fuming Lai}{tencent}
    \icmlauthor{Shaobing Lian}{tencent}
    \icmlauthor{Jie Tang}{tencent}
    \icmlauthor{Yang You}{nus}
  \end{icmlauthorlist}

  \icmlaffiliation{nus}{National University of Singapore}
  \icmlaffiliation{tencent}{Tencent}

  \icmlcorrespondingauthor{Ziming Wu}{jimmyzmwu@tencent.com}
  \icmlcorrespondingauthor{Yang You}{yangyou@nus.edu.sg}

  \icmlkeywords{Speech Language Models, Diffusion Models, Multimodal Learning, Chain-of-Thought Reasoning}

  \vskip 0.3in
]

\printAffiliationsAndNotice{\icmlEqualContribution}

\begin{abstract}
Current speech language models generate responses directly without explicit reasoning, leading to errors that cannot be corrected once audio is produced. We introduce \textbf{``Silent Thought, Spoken Answer''}---a paradigm where speech LLMs generate internal text reasoning alongside spoken responses, with thinking traces informing speech quality. To realize this, we present \method{}, the first diffusion-based speech-text language model supporting both understanding and generation, unifying discrete text and tokenized speech under a single masked diffusion framework. Unlike autoregressive approaches, \method{} jointly generates reasoning traces and speech tokens through iterative denoising, with modality-specific masking schedules. We also construct \dataset{}, the first speech QA dataset with paired text reasoning traces, containing 26K samples totaling 319 hours. Experiments show \method{} achieves state-of-the-art speech-to-speech QA accuracy, outperforming the best baseline by up to 9 points, while attaining the best TTS quality among generative models (6.2\% WER) and preserving language understanding (66.2\% MMLU). Ablations confirm that both the diffusion architecture and thinking traces contribute to these gains.
\end{abstract}

\section{Introduction}
\label{sec:intro}

\begin{figure}[t]
\vskip 0.1in
\begin{center}
\includegraphics[width=0.47\textwidth]{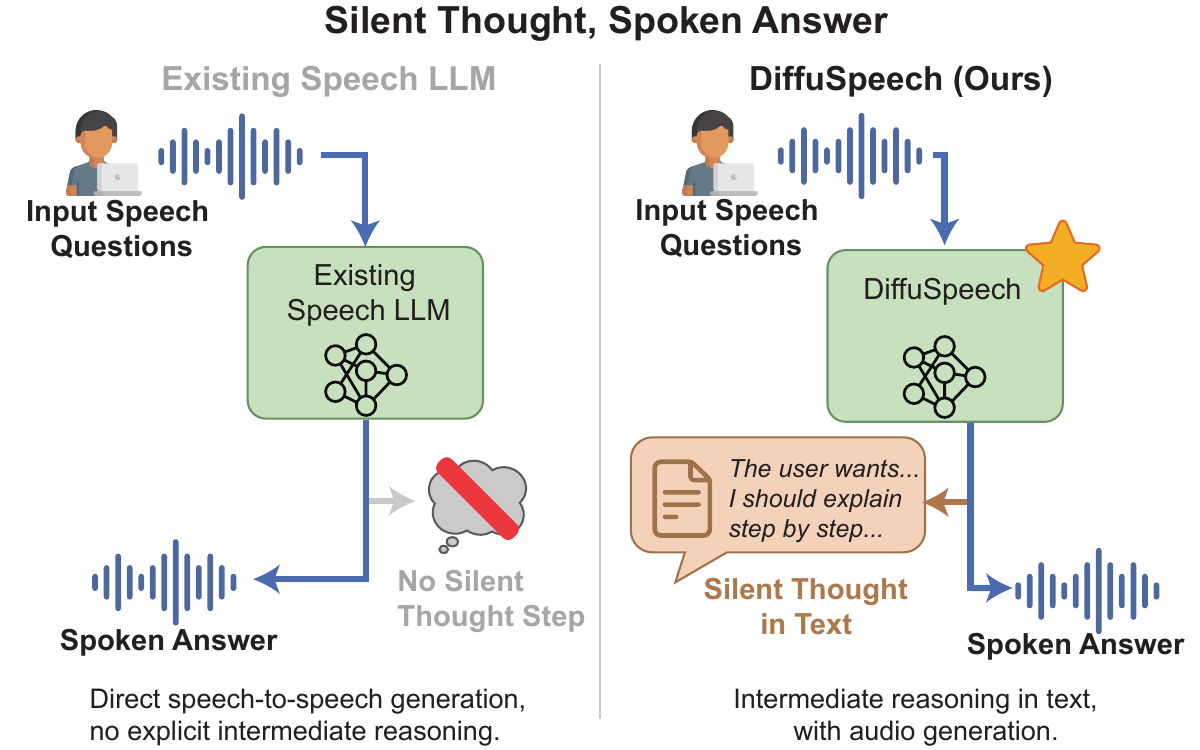}
\caption{The ``Silent Thought, Spoken Answer'' paradigm. Given a spoken question, \method{} jointly generates internal text reasoning (silent thought) alongside the spoken response, with thinking traces informing answer quality. This enables more accurate answers compared to direct speech-to-speech generation.}
\label{fig:teaser}
\end{center}
\vskip -0.2in
\end{figure}

When humans answer complex questions verbally, we think before we speak. This internal reasoning---organizing thoughts, recalling facts, structuring arguments---precedes and shapes our spoken response. Current speech language models lack this capability.  Most existing systems generate speech tokens directly from input, without explicit intermediate reasoning, producing responses that may be fluent but factually incorrect. Crucially, speech generation is \emph{irreversible}: unlike text, where a user can backtrack or edit, once audio is produced, errors cannot be corrected mid-utterance.

Recent advances in speech-text language models~\cite{nguyen2024spiritlm,defossez2024moshi,zhang2023speechgpt} have enabled impressive conversational capabilities, yet they share two fundamental limitations. First, although some recent work explores reasoning capabilities in speech LLMs~\cite{wang2026reasoninggap}, existing systems do not expose \emph{explicit reasoning traces} during generation---the model's thought process remains opaque, preventing users from understanding or verifying how answers are derived. This stands in contrast to text LLMs, where chain-of-thought reasoning has proven highly effective~\cite{wei2023chain,deepseek2025r1}. Second, current speech LLMs are exclusively \emph{autoregressive}, generating tokens left-to-right without bidirectional context---a constraint particularly limiting for speech, where prosody and content are inherently entangled, and where joint modeling of reasoning and response could benefit from global dependencies.

We propose a new paradigm: \textbf{``Silent Thought, Spoken Answer.''} The key insight is to let reasoning inform speech production---the model generates text-based reasoning traces (silent thought) that guide the spoken response. This mirrors human cognition and allows the model to leverage the strong reasoning capabilities of text LLMs while producing natural speech output.

To realize this paradigm, we introduce \textbf{\method{}}, the first diffusion-based speech-text language model supporting both understanding and generation. Unlike autoregressive approaches that generate tokens sequentially, \method{} uses masked diffusion~\cite{sahoo2024simple,nie2025llada} to \emph{jointly} generate both text reasoning and speech tokens through iterative denoising. This unified formulation enables bidirectional context modeling, where thinking traces and speech tokens inform each other during generation. We leverage a unified vocabulary that represents both speech and text as discrete tokens, allowing the model to treat text and audio production as a singular, collaborative process. To further accommodate the different characteristics of each modality, we follow MMaDA~\cite{yang2025mmada} to employ modality-specific masking schedules in our framework.

To facilitate this training paradigm, we construct \textbf{\dataset{}}, the first speech question-answering dataset with paired text reasoning traces. \dataset{} contains 26K samples totaling 319 hours, where each sample includes a spoken question, internal reasoning text, and a spoken answer. The dataset enables research on ``thinking while speaking''---a capability that existing speech LLMs lack.

Extensive experiments show that \method{} achieves state-of-the-art speech-to-speech QA accuracy, outperforming the best baseline (Moshi) by +4.3/+3.0/+9.0 points on LlamaQ/TriviaQA/WebQ. For speech synthesis, \method{} achieves the best TTS quality among all generative models (6.2\% WER on LibriSpeech-Clean). Importantly, it preserves strong language understanding, matching LLaDA on MMLU (66.2\%) while gaining on TriviaQA (+4.7 points). Ablations confirm that both the diffusion architecture and thinking traces contribute to these gains.

Our contributions are:
\begin{itemize}
    \item \textbf{Silent Thought, Spoken Answer}: A new paradigm where speech LLMs jointly generate internal text reasoning alongside spoken responses, enabling reasoning to inform speech quality.
    \item \textbf{First Unified Diffusion Architecture for Speech-Text}: \method{} extends masked diffusion LLMs to the speech-text domain, demonstrating that diffusion is a viable alternative to autoregressive generation for multimodal speech modeling.
    \item \textbf{\dataset{} Dataset}: The first speech QA dataset with text reasoning traces, providing a benchmark and training resource for reasoning-augmented speech generation.
\end{itemize}

\section{Related Work}
\label{sec:related}

\subsection{Speech-Text Language Models}
\label{sec:related_speech}

Recent advances have extended LLMs to process and generate speech. SpiritLM~\cite{nguyen2024spiritlm} pioneered interleaved speech-text modeling with HuBERT tokens, while Moshi~\cite{defossez2024moshi}, SpeechGPT~\cite{zhang2023speechgpt}, Qwen2-Audio~\cite{chu2024qwen2audio}, and MoST~\cite{lou2026most} further advanced conversational and instruction-following capabilities. These models share a common \emph{autoregressive} paradigm, which limits them in two ways: (1) sequential generation prevents parallel decoding, and (2) left-to-right generation cannot leverage bidirectional context for speech. We explore whether \emph{diffusion-based} generation can address these limitations.

\subsection{Diffusion Language Models and Multimodal Extensions}
\label{sec:related_diffusion}

Masked diffusion language models (MDLMs)~\cite{sahoo2024simple,shi2024simplified} formulate text generation as iterative denoising. LLaDA~\cite{nie2025llada} scaled this to 8B parameters with competitive performance, and Dream~\cite{dream2025} improved reasoning capabilities. Several works extend diffusion LLMs to multimodal settings~\cite{yang2025mmada,lladav2025}. Most relevant to our work, DiFFA~\cite{zhou2025diffa} applies diffusion to speech understanding but only supports speech-to-text. \textbf{No diffusion-based LLM supports both speech understanding and generation}---\method{} bridges this gap.

\subsection{Reasoning in Speech-Text Models}
\label{sec:related_reasoning}

Chain-of-thought prompting~\cite{wei2023chain} and explicit reasoning~\cite{deepseek2025r1,yang2025mmada} have proven effective in text and vision-language models. Recent work like TARS~\cite{wang2026reasoninggap} addresses reasoning gaps in speech LLMs but only supports text output. In contrast, our ``Silent Thought, Spoken Answer'' paradigm enables \textbf{text reasoning with speech reply}---jointly generating internal reasoning alongside spoken responses. The \dataset{} dataset provides the first training resource for this capability.


\begin{figure*}[t]
\vskip 0.1in
\begin{center}
\includegraphics[width=0.95\textwidth]{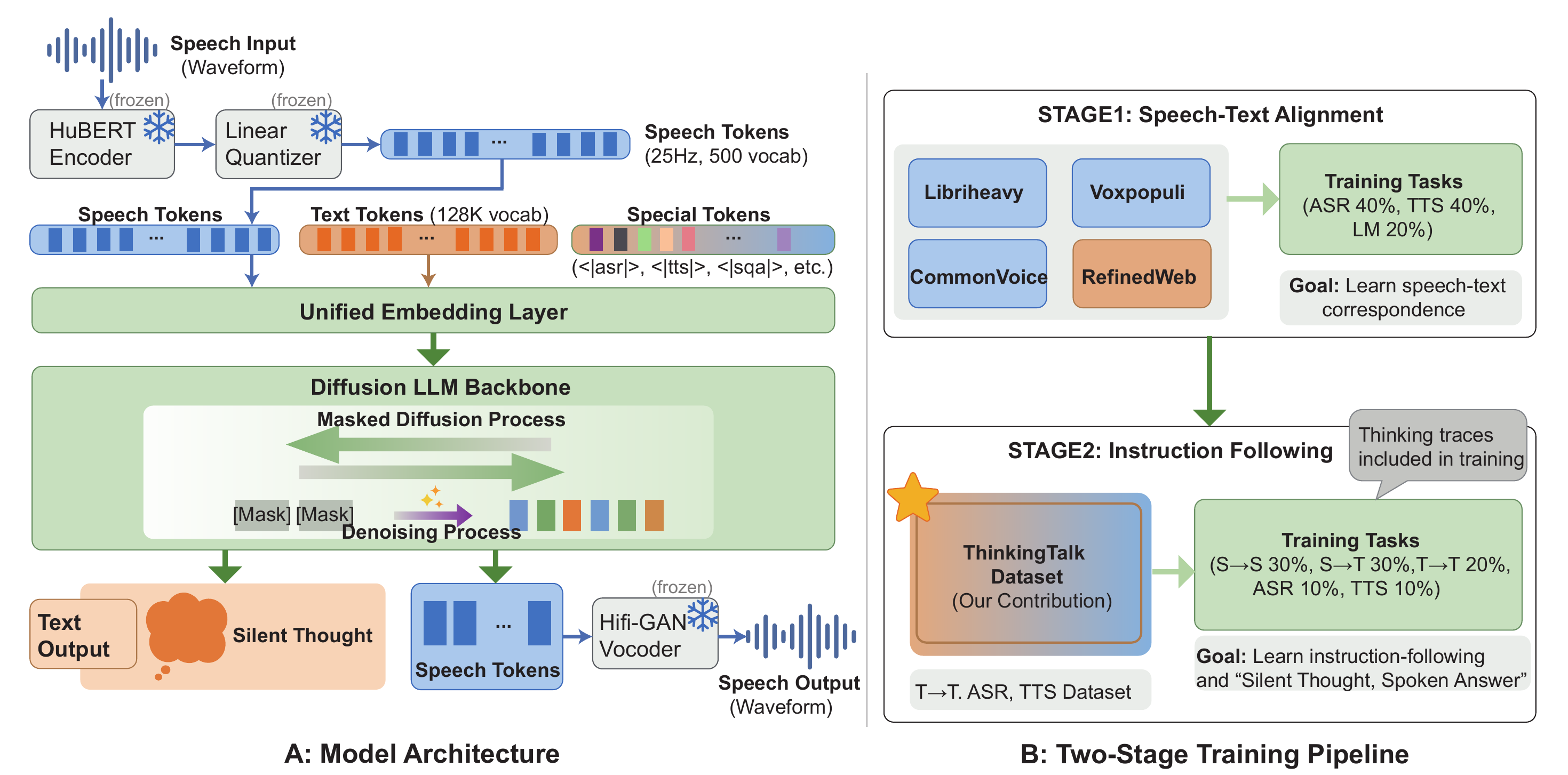}
\caption{\textbf{Overview of \method{} architecture and training pipeline.} \textbf{(a) Model Architecture}: Speech input is encoded by a frozen HuBERT encoder and quantized into discrete tokens, which are combined with text tokens in a unified vocabulary. The LLaDA backbone performs masked diffusion over the combined sequence, generating both text reasoning (silent thought) and speech tokens. Speech output is synthesized by a frozen HiFi-GAN vocoder. \textbf{(b) Two-Stage Training}: Stage 1 performs speech-text alignment using ASR/TTS data from LibriHeavy, VoxPopuli, and CommonVoice, along with text LM data to prevent forgetting. Stage 2 fine-tunes on the \dataset{} dataset for instruction following with thinking traces, enabling the ``Silent Thought, Spoken Answer'' capability.}
\label{fig:architecture}
\end{center}
\vskip -0.2in
\end{figure*}
    
\section{Method}
\label{sec:method}
We present \method{}, a unified diffusion language model for speech and text. \method{} extends masked diffusion language models (MDLMs) to the speech-text multimodal domain, enabling a novel ``Silent Thought, Spoken Answer'' paradigm where the model reasons in text while responding in speech. We first describe our unified representation (\cref{sec:representation}), then review the masked diffusion framework (\cref{sec:preliminaries}), followed by training (\cref{sec:training}) and inference (\cref{sec:inference}). \cref{fig:architecture} provides an overview of the model architecture and training pipeline.

\subsection{Unified Speech-Text Representation}
\label{sec:representation}

We represent both speech and text as discrete tokens in a unified vocabulary $\mathcal{V} = \mathcal{V}_{\text{text}} \cup \mathcal{V}_{\text{special}} \cup \mathcal{V}_{\text{speech}}$.

\paragraph{Speech Tokenization.}
Given raw audio waveform $\bm{a} \in \R^{L}$, we obtain discrete speech tokens via a frozen encoder-quantizer pipeline:
\begin{equation}
\bm{s} = Q(E(\bm{a})) \in \mathcal{V}_{\text{speech}}^{m}
\label{eq:speech_tokenize}
\end{equation}
where $E(\cdot)$ is a HuBERT encoder~\cite{hsu2021hubert} extracting frame-level representations, and $Q(\cdot)$ is a linear quantizer mapping to discrete codes. The encoder operates at a fixed frame rate, producing $m$ tokens for audio of length $L$ (see \cref{app:implementation} for details).

\paragraph{Vocabulary Structure.}
The unified vocabulary extends a pretrained text LLM vocabulary with special tokens $\mathcal{V}_{\text{special}}$ for task control and speech tokens $\mathcal{V}_{\text{speech}}$. Speech token IDs are offset by $|\mathcal{V}_{\text{text}}| + |\mathcal{V}_{\text{special}}|$ to avoid collision. The full vocabulary specification is provided in \cref{app:implementation}.

\paragraph{Sequence Format.}
For multimodal tasks, we structure sequences as $\bm{x} = [\tau, \bm{c}, \bm{y}]$ where $\tau \in \mathcal{V}_{\text{special}}$ is a task token, $\bm{c}$ is the conditioning input, and $\bm{y}$ is the target output. For our key task---speech-to-speech with reasoning (S2S)---the format is:
\begin{equation}
\bm{x} = [\tau_{\text{s2s}}, \underbrace{\bm{s}_{\text{user}}}_{\text{input speech}}, \underbrace{\bm{t}_{\text{think}}}_{\text{reasoning (text)}}, \underbrace{\bm{s}_{\text{reply}}}_{\text{output speech}}]
\label{eq:s2s_format}
\end{equation}
where delimiters (omitted for clarity) separate each component. This unified format enables the model to leverage text reasoning $\bm{t}_{\text{think}}$ while generating spoken responses $\bm{s}_{\text{reply}}$. Formats for other tasks (TTS, ASR, S2T, T2T) follow analogously (see \cref{app:implementation}).

\subsection{Masked Diffusion for Speech and Text}
\label{sec:preliminaries}

Given the unified vocabulary $\mathcal{V}$, we apply Masked Diffusion Language Models (MDLMs)~\cite{sahoo2024simple,nie2025llada} to jointly model speech and text. Given a token sequence $\bm{x}_0 = (x_1, \ldots, x_n) \in \mathcal{V}^n$, the forward process $q$ corrupts it by independently masking each token with probability $\gamma(t)$ at noise level $t \in [0, 1]$:
\begin{equation}
q(\bm{x}_t | \bm{x}_0) = \prod_{i=1}^{n} \left[ \gamma(t) \cdot \mathbf{1}_{[x_t^{(i)} = \texttt{m}]} + (1-\gamma(t)) \cdot \mathbf{1}_{[x_t^{(i)} = x_0^{(i)}]} \right]
\label{eq:forward}
\end{equation}
where $\texttt{m} \in \mathcal{V}$ denotes the mask token. The reverse process learns to predict $\bm{x}_0$ from $\bm{x}_t$ via a transformer $f_\theta$:
\begin{equation}
p_\theta(x_0^{(i)} | \bm{x}_t) = \text{softmax}\left(f_\theta(\bm{x}_t)_i\right), \quad \forall i : x_t^{(i)} = \texttt{m}
\label{eq:reverse}
\end{equation}
The training objective minimizes cross-entropy on masked positions:
\begin{equation}
\mathcal{L}_{\text{MDLM}} = \E_{t, \bm{x}_t} \left[ \sum_{i \in \mathcal{M}_t} -\log p_\theta(x_0^{(i)} | \bm{x}_t) \right]
\label{eq:mdlm_loss}
\end{equation}
where $\mathcal{M}_t = \{i : x_t^{(i)} = \texttt{m}\}$ is the set of masked positions at noise level $t$. We use the cosine masking schedule:
\begin{equation}
\gamma(t) = \cos\left(\frac{\pi}{2} \cdot (1-t)\right)
\label{eq:cosine_schedule}
\end{equation}
which provides smooth transitions from full masking ($t=1$) to no masking ($t=0$).

\subsection{Multi-Task Training}
\label{sec:training}

\method{} is trained via a two-stage curriculum (\cref{fig:architecture}b). \textbf{Stage 1} performs speech-text alignment with ASR (speech$\to$text), TTS (text$\to$speech), and text LM tasks. \textbf{Stage 2} enables instruction following with S2S (speech$\to$thinking+speech), S2T (speech$\to$thinking+text), T2T (text$\to$thinking+text), plus continued ASR/TTS. Let $\mathcal{T} = \{k_1, \ldots, k_K\}$ denote the set of tasks, each with sampling probability $p_k$ and loss coefficient $\lambda_k$.

\paragraph{Selective Masking.}
For each sample $\bm{x} = [\tau, \bm{c}, \bm{y}]$ of task $k$, we apply masking \emph{only} to the target $\bm{y}$ while keeping the condition $\bm{c}$ intact. Specifically, we partition positions into condition set $\mathcal{C}$ and target set $\mathcal{Y}$, then apply the forward process (\cref{eq:forward}) only to $\mathcal{Y}$:
\begin{equation}
x_t^{(i)} = 
\begin{cases}
x_0^{(i)} & \text{if } i \in \mathcal{C} \\
\texttt{m} \text{ w.p. } \gamma(t), \text{ else } x_0^{(i)} & \text{if } i \in \mathcal{Y}
\end{cases}
\label{eq:selective_mask}
\end{equation}
This ensures the model learns the correct conditional generation: $p(\bm{y} | \bm{c}, \tau)$.

\paragraph{Training Objective.}
At each step, we sample task $k \sim \text{Categorical}(\{p_k\})$, draw a sample $\bm{x}^{(k)}$, and compute the loss on masked target positions:
\begin{equation}
\mathcal{L} = \sum_{k \in \mathcal{T}} \lambda_k \cdot \E_{\bm{x}^{(k)}, t} \left[ \sum_{i \in \mathcal{M}_t \cap \mathcal{Y}_k} -\log p_\theta(x_0^{(i)} | \bm{x}_t^{(k)}) \right]
\label{eq:total_loss}
\end{equation}
The two-stage curriculum and task proportions $\{p_k\}$ are detailed in \cref{app:implementation}.

\subsection{Inference: Silent Thought, Spoken Answer}
\label{sec:inference}

At inference, \method{} generates both reasoning and response in a single diffusion process, which we term ``Silent Thought, Spoken Answer.'' Given conditioning input $\bm{c}$, we initialize the target with all mask tokens and iteratively denoise over $T$ steps.

\paragraph{Iterative Denoising.}
Let $n = |\mathcal{Y}|$ be the number of target positions. At each step $i \in \{T, T-1, \ldots, 1\}$, we: (1) compute predictions $p_\theta(\cdot | \bm{x}_i)$ for all masked positions, (2) select the top-$k_i$ most confident predictions to unmask, where $k_i = \lceil n \cdot (T-i+1)/T \rceil$ follows a linear schedule, and (3) update $\bm{x}_{i-1}$ by replacing selected masks with predicted tokens. This confidence-based unmasking allows the model to commit to easy decisions first and refine harder ones later. The full inference procedure is given in \cref{alg:inference}.


\begin{algorithm}[t]
\caption{Inference: Silent Thought, Spoken Answer}
\label{alg:inference}
\begin{algorithmic}
\STATE {\bfseries Input:} Condition $\bm{c}$, task token $\tau$, target length $n$, steps $T$
\STATE {\bfseries Output:} Generated target $\hat{\bm{y}} = [\hat{\bm{t}}_{\text{think}}, \hat{\bm{s}}_{\text{reply}}]$
\STATE Initialize: $\bm{x}_T \leftarrow [\tau, \bm{c}, \underbrace{\texttt{m}, \ldots, \texttt{m}}_{n}]$
\FOR{$i = T, T-1, \ldots, 1$}
    \STATE Compute predictions: $\bm{p} = p_\theta(\cdot | \bm{x}_i)$
    \STATE Compute confidence: $\text{conf}_j = \max_{v \in \mathcal{V}} p_j(v)$ for masked $j$ \hfill \textit{// $\mathcal{V}$: vocabulary}
    \STATE Determine tokens to unmask: $k_i = \lceil n \cdot (T-i+1)/T \rceil$
    \STATE Select top-$k_i$ positions $\mathcal{S}_i$ by confidence
    \FOR{$j \in \mathcal{S}_i$}
        \STATE $x_{i-1}^{(j)} \leftarrow \arg\max_{v \in \mathcal{V}} p_j(v)$
    \ENDFOR
\ENDFOR
\STATE Extract $\hat{\bm{y}}$ from $\bm{x}_0$
\STATE {\bfseries Return:} $\hat{\bm{t}}_{\text{think}}$, $\text{Vocoder}(\hat{\bm{s}}_{\text{reply}})$
\end{algorithmic}
\end{algorithm}

\section{\dataset{} Dataset}
\label{sec:dataset}

To enable the ``Silent Thought, Spoken Answer'' paradigm, we construct \dataset{}, the first speech question-answering dataset with paired text reasoning traces. Unlike existing speech QA datasets that only contain spoken questions and answers, \dataset{} provides explicit \emph{thinking traces}---internal reasoning text that bridges the question understanding and answer generation. This section describes the construction pipeline (\cref{sec:pipeline}) and dataset characteristics (\cref{sec:statistics}).

\subsection{Construction Pipeline}
\label{sec:pipeline}

\begin{figure}[t]
\vskip 0.1in
\begin{center}
\includegraphics[width=\columnwidth]{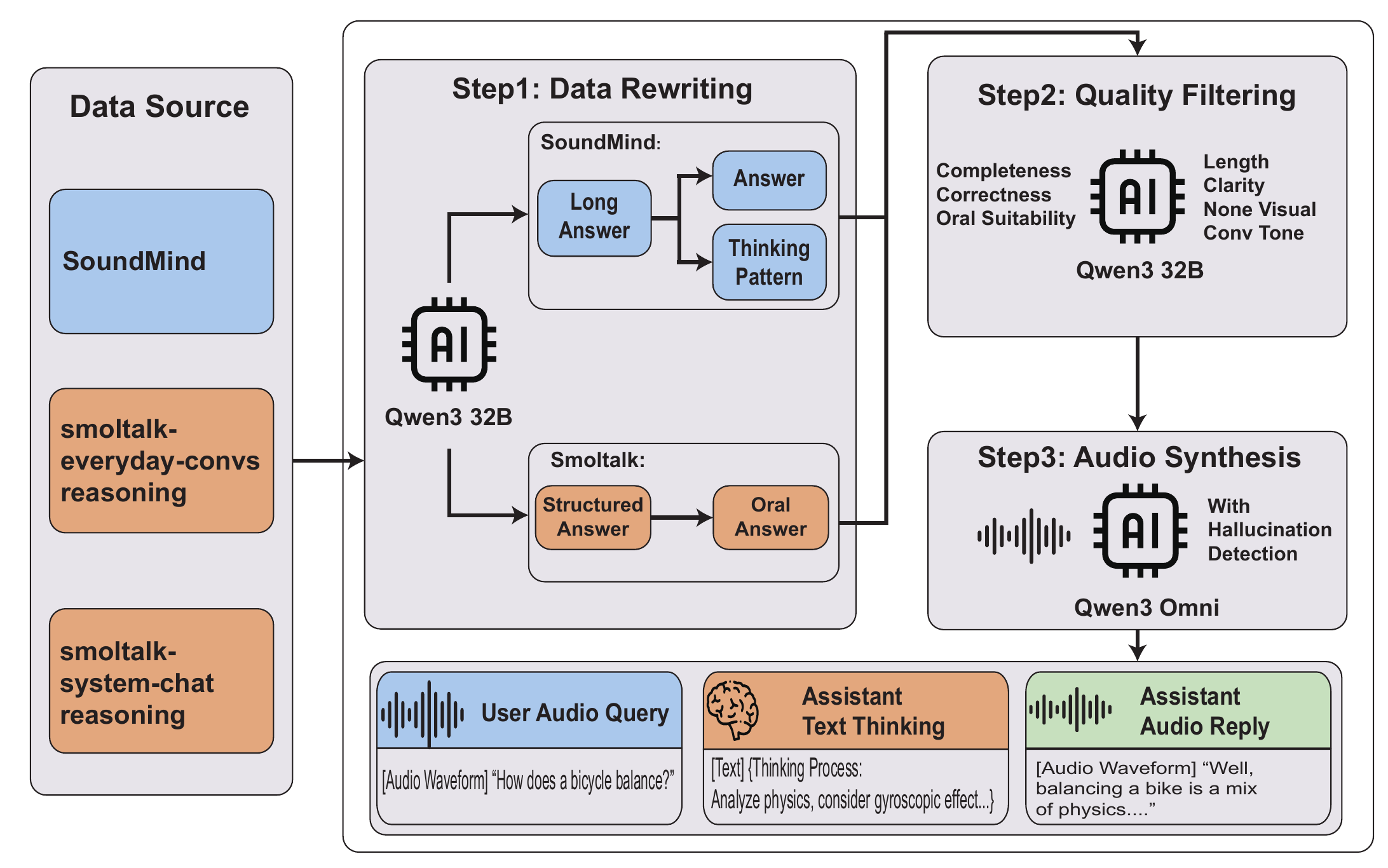}
\caption{Construction pipeline for the \dataset{} dataset. We rewrite existing text QA datasets into an oral style with explicit thinking traces, filter for speech suitability using an LLM judge, and synthesize high-quality audio for both user questions and assistant responses.}
\label{fig:pipeline}
\end{center}
\vskip -0.2in
\end{figure}

\dataset{} is constructed through a three-stage pipeline illustrated in \cref{fig:pipeline}: (1) data rewriting, (2) quality filtering, and (3) audio synthesis.

\paragraph{Stage 1: Data Rewriting.}
We source from two text QA datasets: Smoltalk2~\cite{allal2025smollm2} (conversational QA) and SoundMind~\cite{diao2025soundmind} (logical reasoning QA). Raw samples are rewritten using Qwen-32B~\cite{qwen2025qwen3} to produce three components: (1) \textbf{user input} in natural spoken style, (2) \textbf{assistant thinking} containing step-by-step reasoning, and (3) \textbf{assistant reply} as a natural spoken response. The rewriting ensures content is suitable for audio---avoiding visual references, mathematical notation, and code snippets that cannot be naturally verbalized.

\paragraph{Stage 2: Quality Filtering.}
We employ an LLM-based quality judge (Qwen3-32B) to filter samples across seven dimensions: completeness, correctness, oral suitability, clarity, length appropriateness, visual independence, and conversational tone. Each dimension is scored 1--5, with weighted aggregation (emphasizing correctness and oral suitability). Samples with overall score $< 3.5$ or any critical dimension (correctness, oral suitability, visual independence) scoring $< 3$ are rejected. This stage reduces the dataset from 32,292 to 26,387 samples (81.7\% retention).

\paragraph{Stage 3: Audio Synthesis.}
User inputs are synthesized using MegaTTS3~\cite{jiang2025megatts3} with 4 diverse reference speakers from LibriSpeech~\cite{panayotov2015librispeech} to simulate varied user voices. Assistant replies are synthesized using Qwen3-Omni~\cite{xu2025qwen25omni} with a single consistent voice for coherent assistant identity. We apply hallucination detection via words-per-second (WPS) validation, rejecting audio with abnormal speaking rates ($< 1.5$ or $> 5.5$ WPS).

\subsection{Dataset Statistics}
\label{sec:statistics}

\paragraph{Scale and Distribution.}
\cref{tab:dataset_stats} summarizes \dataset{} statistics. The dataset contains \textbf{26,387 samples} totaling \textbf{319 hours}, with thinking traces averaging 77.5 words and spoken replies averaging 102 words. The three sources contribute complementary content: Smoltalk-SystemChat (88\%) provides practical knowledge, Smoltalk-EverydayConv (6\%) covers casual topics, and SoundMind (6\%) contributes logical reasoning tasks.

\begin{table}[t]
\caption{Statistics of the \dataset{} dataset.}
\label{tab:dataset_stats}
\vskip 0.1in
\begin{center}
\begin{small}
\begin{tabular}{lc}
\toprule
\textbf{Statistic} & \textbf{Value} \\
\midrule
Total samples & 26,387 \\
Total audio duration & 319 hours \\
Mean sample duration & 43.5 sec \\
\midrule
\multicolumn{2}{l}{\textit{Thinking Trace (Text)}} \\
\quad Mean length & 77.5 words \\
\quad Median length & 75 words \\
\midrule
\multicolumn{2}{l}{\textit{Assistant Reply (Speech)}} \\
\quad Mean length & 102 words \\
\quad Median length & 100 words \\
\midrule
\multicolumn{2}{l}{\textit{Source Distribution}} \\
\quad Smoltalk-SystemChat & 23,247 (88.1\%) \\
\quad Smoltalk-EverydayConv & 1,673 (6.3\%) \\
\quad SoundMind & 1,467 (5.6\%) \\
\bottomrule
\end{tabular}
\end{small}
\end{center}
\vskip -0.1in
\end{table}

\paragraph{Unique Properties.}
Unlike prior speech datasets, \dataset{} explicitly captures the \emph{reasoning process} that connects questions to answers. This enables training models that can ``think while speaking''---a capability that existing speech LLMs lack. The thinking traces are designed to be silent (text-only) during inference, serving as internal reasoning that improves answer quality without being vocalized. Full dataset statistics and example samples are provided in \cref{app:dataset}.

\section{Experiments}
\label{sec:experiments}

We evaluate \method{} on three categories of tasks: language understanding, speech processing (ASR/TTS), and speech question-answering. Our experiments address three questions: (1) Does speech training preserve language capabilities? (2) Can diffusion models achieve competitive ASR/TTS? (3) Does ``Silent Thought'' improve spoken responses?

\subsection{Experimental Setup}
\label{sec:setup}

\paragraph{Baselines.}
We compare against three categories of models: (1) \textbf{Autoregressive speech LLMs}: Phi-4-Multimodal~\cite{microsoft2025phi4}, MinMo~\cite{chen2025minmo}, Moshi~\cite{defossez2024moshi}, SpiritLM~\cite{nguyen2024spiritlm}, Llama-Omni2~\cite{fang2025llamaomni2}, Qwen2-Audio~\cite{chu2024qwen2audio}, and SpeechGPT~\cite{zhang2023speechgpt}; (2) \textbf{Diffusion speech LLM}: DiFFA~\cite{zhou2025diffa} (speech-to-text only); (3) \textbf{Diffusion text LLM}: LLaDA~\cite{nie2025llada} (our base model). Note that some baselines support only S$\to$T (no speech generation).

\paragraph{Evaluation Benchmarks.}
\textit{Language Understanding}: MMLU~\cite{hendrycks2021mmlu}, TriviaQA~\cite{joshi2017triviaqa}, GSM8K~\cite{cobbe2021gsm8k}.
\textit{Speech Processing}: LibriSpeech~\cite{panayotov2015librispeech} (LS-Clean, LS-Other), VoxPopuli~\cite{wang2021voxpopuli}, CommonVoice~\cite{ardila2020commonvoice} for ASR (WER$\downarrow$) and TTS (WER$\downarrow$ via Whisper-large-v3~\cite{radford2023whisper} on generated speech).
\textit{Speech QA}: LlamaQuestions~\cite{nachmani2024llamaquestions}, TriviaQA (spoken), WebQuestions~\cite{berant2013webq}, and OpenBookQA~\cite{mihaylov2018obqa} with both S$\to$T and S$\to$S evaluation.
\textit{Speech Reasoning}: MMSU~\cite{wang2025mmsu} evaluated via VoiceBench~\cite{chen2024voicebench}, a massive multi-task spoken language understanding benchmark (S$\to$T only).

\subsection{Main Results}
\label{sec:results}

\begin{table*}[t]
\caption{Speech question-answering results (Accuracy\%$\uparrow$). S$\to$T Avg is computed over all four benchmarks; S$\to$S Avg over LlamaQ, TriviaQA, and WebQ (OBQA is S$\to$T only). \xmark/\cmark{} indicates S$\to$S capability. \gray{``/''} = unsupported. Improvements over best baseline shown with {\footnotesize\textcolor{impgreen}{$\uparrow$}}.}
\label{tab:speech_qa}
\vskip 0.1in
\begin{center}
\begin{footnotesize}
\renewcommand{\arraystretch}{1.15}
\setlength{\tabcolsep}{4pt}
\begin{tabular}{@{}l@{\hspace{4pt}}c@{\hspace{3pt}}c@{\hspace{6pt}}cc@{\hspace{5pt}}cc@{\hspace{5pt}}cc@{\hspace{5pt}}c@{\hspace{8pt}}c@{\hspace{6pt}}c@{}}
\toprule
& & & \multicolumn{2}{c}{\textbf{LlamaQ}} & \multicolumn{2}{c}{\textbf{TriviaQA}} & \multicolumn{2}{c}{\textbf{WebQ}} & \textbf{OBQA} & & \\
\cmidrule(lr){4-5} \cmidrule(lr){6-7} \cmidrule(lr){8-9} \cmidrule(lr){10-10}
\textbf{Model} & \textbf{Type} & \textbf{S$\to$S} & S$\to$T & S$\to$S & S$\to$T & S$\to$S & S$\to$T & S$\to$S & S$\to$T & \textbf{S$\to$T Avg} & \textbf{S$\to$S Avg} \\
\midrule
\rowcolor{tablerow1}
Qwen2-Audio & AR & \xmark & 55.4 & \gray{/} & 38.5 & \gray{/} & 41.2 & \gray{/} & 49.5 & 46.2 & \gray{/} \\
\rowcolor{tablerow2}
Phi-4-Multimodal & AR & \xmark & 60.2 & \gray{/} & 22.8 & \gray{/} & 26.6 & \gray{/} & \textbf{65.9} & 43.9 & \gray{/} \\
\midrule
\rowcolor{tablerow1}
SpeechGPT & AR & \cmark & 21.6 & 5.4 & 14.8 & 8.2 & 6.3 & 5.5 & 12.4 & 13.8 & 6.4 \\
\rowcolor{tablerow2}
SpiritLM & AR & \cmark & 40.1 & 38.7 & 6.5 & 4.2 & 12.3 & 9.6 & 21.7 & 20.2 & 17.5 \\
\rowcolor{tablerow1}
Moshi & AR & \cmark & 68.3 & 64.2 & 42.1 & 30.5 & 47.5 & 40.7 & 26.0 & 46.0 & 45.1 \\
\rowcolor{tablerow2}
MinMo & AR & \cmark & \textbf{76.5} & 63.8 & 40.1 & 25.5 & 55.0 & 39.9 & 44.5 & 54.0 & 43.1 \\
\rowcolor{tablerow1}
Llama-Omni2 & AR & \cmark & 70.3 & 60.7 & 35.3 & 23.9 & 40.4 & 37.1 & 28.1 & 43.5 & 40.6 \\
\midrule
\rowcolor{tablerow2}
DiFFA & Diff. & \xmark & 58.3 & \gray{/} & 36.0 & \gray{/} & 43.4 & \gray{/} & 35.6 & 43.3 & \gray{/} \\
\ours
\method{} & Diff. & \cmark & 72.0 & \imp{68.5}{4.3} & \imp{45.4}{3.3} & \imp{33.5}{3.0} & \imp{61.5}{6.5} & \imp{49.7}{9.0} & 51.3 & \imp{57.6}{3.6} & \imp{50.6}{5.5} \\
\bottomrule
\end{tabular}
\end{footnotesize}
\end{center}
\vskip -0.1in
\end{table*}

\paragraph{Speech Question-Answering.}
\cref{tab:speech_qa} presents results on four speech QA benchmarks, our primary evaluation for the ``Silent Thought, Spoken Answer'' paradigm. We evaluate both S$\to$T (speech question, text answer) and S$\to$S (speech question, speech answer) modes where applicable.

\begin{figure}[t]
\vskip 0.1in
\begin{center}
\includegraphics[width=0.9\columnwidth]{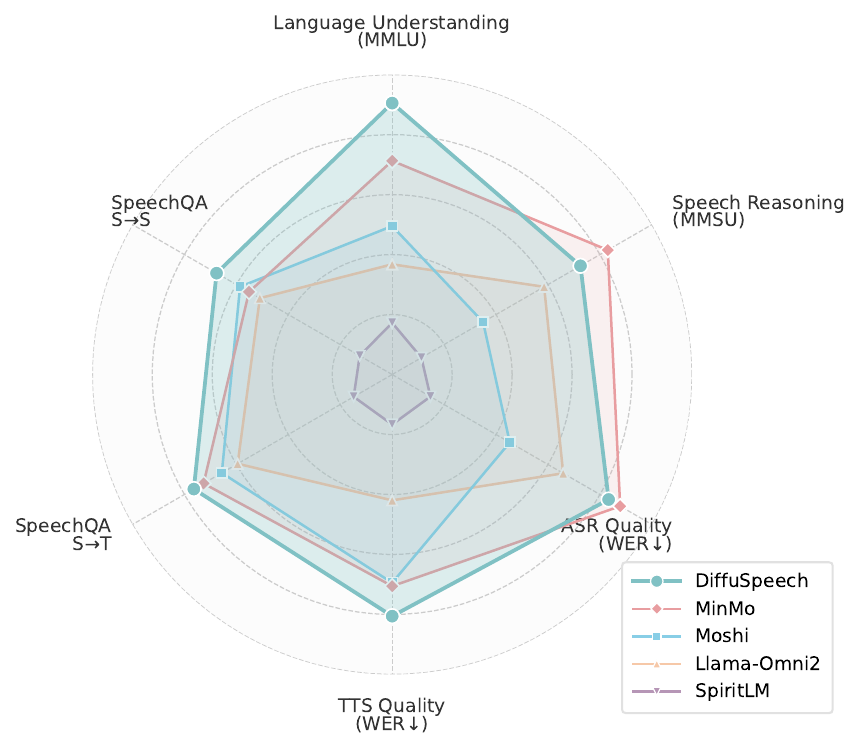}
\caption{Capability comparison across six dimensions, including speech reasoning (MMSU). \method{} achieves well-rounded performance, excelling in Speech QA while maintaining strong language understanding and competitive speech reasoning. Values are normalized to [0, 100] with higher being better (WER metrics are inverted).}
\label{fig:radar_chart}
\end{center}
\vskip -0.2in
\end{figure}

\method{} achieves the \textbf{best S$\to$S performance} across all benchmarks with S$\to$S evaluation: 68.5\% on LlamaQ (+4.3 over Moshi), 33.5\% on TriviaQA (+3.0), and 49.7\% on WebQ (+9.0). Notably, while most models show significant degradation from S$\to$T to S$\to$S (e.g., MinMo drops 12.7/14.6/15.1 points across benchmarks), \method{} maintains strong S$\to$S performance with smaller gaps. On average, \method{} achieves 57.6\% S$\to$T accuracy (+3.6 over MinMo) and 50.6\% S$\to$S accuracy (+5.5 over Moshi), demonstrating that jointly generating reasoning traces alongside speech enables more accurate spoken responses. Additional speech reasoning results on MMSU are provided in \cref{fig:radar_chart} and \cref{app:mmsu_results}.


\paragraph{Speech Processing.}
\cref{tab:speech_processing} presents ASR and TTS results. For ASR, \method{} achieves competitive WER (3.0\% on LS-Clean, 7.1\% on VoxPopuli), comparable to specialized models like MinMo and Qwen2-Audio. For TTS, \method{} achieves the \textbf{best WER among all speech-generative models} across all benchmarks (6.2\% on LS-Clean, 10.3\% on VoxPopuli), outperforming MinMo (6.7\%, 10.9\%) and Moshi (7.0\%, 10.6\%). This demonstrates that diffusion-based generation produces high-quality, intelligible speech.

\begin{table}[t]
\caption{Speech processing results (WER\%$\downarrow$). \gray{``/''} indicates unsupported (no speech generation). Full results including CommonVoice in \cref{app:full_results}.}
\label{tab:speech_processing}
\vskip 0.1in
\begin{center}
\begin{small}
\renewcommand{\arraystretch}{1.12}
\setlength{\tabcolsep}{3pt}
\begin{tabular}{@{}l@{\hspace{4pt}}cc@{\hspace{4pt}}cc@{\hspace{4pt}}cc@{}}
\toprule
& \multicolumn{2}{c}{\textbf{LS-Clean}} & \multicolumn{2}{c}{\textbf{LS-Other}} & \multicolumn{2}{c}{\textbf{VoxPopuli}} \\
\cmidrule(lr){2-3} \cmidrule(lr){4-5} \cmidrule(lr){6-7}
\textbf{Model} & ASR & TTS & ASR & TTS & ASR & TTS \\
\midrule
\rowcolor{tablerow1}
Qwen2-Audio & \textbf{1.8} & \gray{/} & \textbf{3.5} & \gray{/} & 7.1 & \gray{/} \\
\rowcolor{tablerow2}
Phi-4-Multimodal & 2.1 & \gray{/} & 3.6 & \gray{/} & \textbf{6.3} & \gray{/} \\
\midrule
\rowcolor{tablerow1}
SpeechGPT & 11.0 & 14.1 & 16.7 & 15.3 & 18.2 & 21.3 \\
\rowcolor{tablerow2}
SpiritLM & 6.0 & 6.7 & 11.0 & 9.5 & 14.3 & 19.4 \\
\rowcolor{tablerow1}
Moshi & 5.5 & 7.0 & 12.0 & 7.2 & 8.8 & 10.6 \\
\rowcolor{tablerow2}
MinMo & \textbf{1.8} & 6.7 & 3.9 & 7.5 & 6.7 & 10.9 \\
\rowcolor{tablerow1}
Llama-Omni2 & 3.5 & 10.1 & 4.0 & 9.2 & 9.5 & 12.4 \\
\midrule
\ours
\method{} & 3.0 & \textbf{6.2} & 4.3 & \textbf{6.8} & 7.1 & \textbf{10.3} \\
\bottomrule
\end{tabular}
\end{small}
\end{center}
\vskip -0.1in
\end{table}

\paragraph{Language Understanding.}
\cref{tab:lm_results} shows that \method{} preserves strong language capabilities after speech training. Compared to LLaDA (our base model), \method{} achieves slightly higher scores on all three benchmarks (+0.3 MMLU, +4.7 TriviaQA, +2.5 GSM8K), suggesting that multi-task speech-text training may even benefit text understanding. Among speech-capable models, \method{} also achieves competitive performance across three different task types.

\begin{table}[t]
\caption{Language understanding results. All models are evaluated in text-in, text-out mode. Improvements over LLaDA (base model) shown with {\footnotesize\textcolor{impgreen}{$\uparrow$}}.}
\label{tab:lm_results}
\vskip 0.1in
\begin{center}
\begin{small}
\renewcommand{\arraystretch}{1.15}
\begin{tabular}{@{}l@{\hspace{8pt}}c@{\hspace{12pt}}c@{\hspace{12pt}}c@{\hspace{12pt}}c@{}}
\toprule
\textbf{Model} & \textbf{Type} & \textbf{MMLU} & \textbf{TriviaQA} & \textbf{GSM8K} \\
\midrule
\rowcolor{tablerow1}
Phi-4-Multimodal & AR & \textbf{67.3} & 58.5 & \textbf{88.6} \\
\rowcolor{tablerow2}
MinMo & AR & 58.5 & 54.8 & 58.1 \\
\rowcolor{tablerow1}
Moshi & AR & 49.8 & 48.5 & 40.3 \\
\rowcolor{tablerow2}
SpiritLM & AR & 36.9 & 42.0 & 21.5 \\
\midrule
\rowcolor{tablerow1}
LLaDA & Diff. & 65.9 & 55.6 & 70.3 \\
\ours
\method{} & Diff. & \imp{66.2}{0.3} & \imp{60.3}{4.7} & \imp{72.8}{2.5} \\
\bottomrule
\end{tabular}
\end{small}
\end{center}
\vskip -0.1in
\end{table}

\subsection{Ablation Studies}
\label{sec:ablation}

We conduct three ablation studies to validate our design choices: (1) AR vs.\ diffusion for speech adaptation, (2) effect of thinking traces, and (3) sample efficiency of diffusion generation.

\begin{figure}[t]
\vskip 0.1in
\begin{center}
\includegraphics[width=\columnwidth]{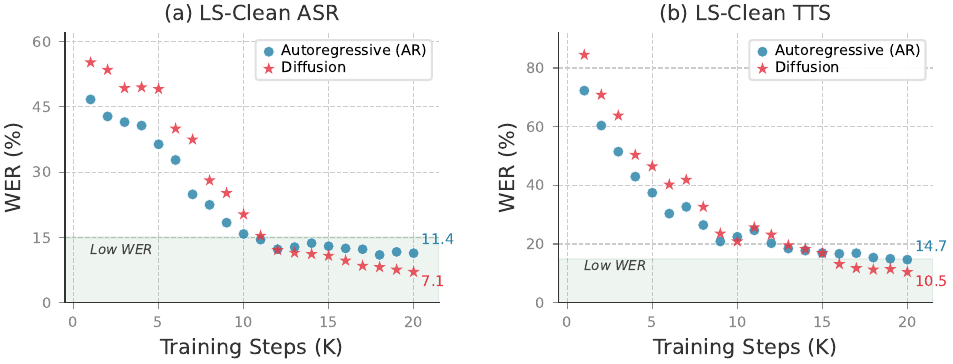}
\caption{Training dynamics of AR vs.\ Diffusion for speech adaptation. \textbf{(a)} ASR and \textbf{(b)} TTS on LibriSpeech-Clean. While AR (blue) shows faster initial learning, Diffusion (red) surpasses AR mid-training and achieves lower final WER, demonstrating better speech adaptation potential.}
\label{fig:ablation_arch}
\end{center}
\vskip -0.2in
\end{figure}

\paragraph{AR vs.\ Diffusion for Speech Adaptation.}
We compare autoregressive and diffusion paradigms for adapting LLMs to speech. Starting from base LLMs with similar language capabilities (Llama-3.1-8B and LLaDA-8B), we apply identical speech alignment training (Stage 1) and track WER throughout training. \cref{fig:ablation_arch} shows the training dynamics on LibriSpeech-Clean for both ASR and TTS tasks.

AR models show faster initial improvement in the first few thousand steps. However, \textbf{diffusion surpasses AR} at around 10K steps for ASR and 15K steps for TTS, ultimately achieving significantly lower final WER (7.1\% vs.\ 11.4\% for ASR, 10.5\% vs.\ 14.7\% for TTS). This suggests that while AR adapts quickly to surface patterns, diffusion's bidirectional modeling enables better long-term learning of speech-text alignment, demonstrating stronger potential for speech adaptation.

\paragraph{Effect of Thinking Traces.}
We ablate the impact of thinking traces by fine-tuning models on \dataset{} with and without the text reasoning component. We compare two base models: SpiritLM (AR) and \method{} Stage 1 (Diffusion). Both models are fine-tuned on \dataset{} either with full samples (including thinking traces) or with thinking traces removed.

\cref{tab:ablation_thinking} shows S$\to$S accuracy on speech QA tasks. Adding thinking traces improves both models substantially, but \method{} shows consistently larger gains: +13.4 points average compared to +10.5 for SpiritLM. This validates that (1) ``Silent Thought'' is effective across architectures, and (2) diffusion models can better leverage reasoning traces, possibly due to the joint generation of thinking and speech tokens.

\begin{table}[t]
\caption{Effect of thinking traces on S$\to$S accuracy (\%$\uparrow$). $\Delta$: improvement from adding thinking traces.}
\label{tab:ablation_thinking}
\vskip 0.1in
\begin{center}
\begin{small}
\renewcommand{\arraystretch}{1.12}
\setlength{\tabcolsep}{4pt}
\begin{tabular}{@{}lccc|ccc@{}}
\toprule
& \multicolumn{3}{c|}{\textbf{SpiritLM (AR)}} & \multicolumn{3}{c}{\textbf{\method{} (Diff.)}} \\
\cmidrule(lr){2-4} \cmidrule(lr){5-7}
\textbf{Dataset} & w/o & w/ & $\Delta$ & w/o & w/ & $\Delta$ \\
\midrule
\rowcolor{tablerow1}
LlamaQ & 44.9 & 57.4 & \textcolor{impgreen}{+12.5} & 53.1 & 68.5 & \textcolor{impgreen}{\textbf{+15.4}} \\
\rowcolor{tablerow2}
TriviaQA & 18.3 & 26.1 & \textcolor{impgreen}{+7.8} & 21.4 & 33.5 & \textcolor{impgreen}{\textbf{+12.1}} \\
\rowcolor{tablerow1}
WebQ & 21.5 & 32.8 & \textcolor{impgreen}{+11.3} & 36.9 & 49.7 & \textcolor{impgreen}{\textbf{+12.8}} \\
\midrule
\rowcolor{tablerow2}
\textit{Average} & 28.2 & 38.8 & \textcolor{impgreen}{+10.5} & 37.1 & 50.6 & \textcolor{impgreen}{\textbf{+13.4}} \\
\bottomrule
\end{tabular}
\end{small}
\end{center}
\vskip -0.1in
\end{table}


\paragraph{Sample Efficiency.}
A key advantage of diffusion models over autoregressive models is the ability to generate multiple tokens in parallel. We study this by varying the number of denoising steps while keeping the target sequence length fixed at 1024 tokens. \cref{tab:ablation_efficiency} shows that \method{} maintains strong performance even with significantly fewer steps: at 512 steps (2$\times$ speedup), TTS quality actually \emph{improves} (6.20\% vs.\ 8.45\% WER) while SpeechQA remains stable (72.06\% vs.\ 72.13\%). Even at 256 steps (4$\times$ speedup), performance degrades gracefully (7.60\% TTS WER, 70.52\% SpeechQA). These results underscore the efficiency potential of diffusion-based multimodal language models compared with AR approaches, where generation cost scales linearly with sequence length.

\begin{table}[t]
\caption{Sample efficiency ablation. Performance vs.\ denoising steps with 1024 target tokens. Fewer steps enable faster inference.}
\label{tab:ablation_efficiency}
\vskip 0.1in
\begin{center}
\begin{small}
\renewcommand{\arraystretch}{1.12}
\setlength{\tabcolsep}{6pt}
\begin{tabular}{@{}c@{\hspace{12pt}}c@{\hspace{12pt}}c@{\hspace{12pt}}c@{}}
\toprule
\textbf{Steps} & \textbf{TTS} (WER$\downarrow$) & \textbf{ASR} (WER$\downarrow$) & \textbf{SpeechQA} (Acc$\uparrow$) \\
\midrule
\rowcolor{tablerow1}
1024 & 8.45 & \textbf{3.08} & \textbf{72.13} \\
\rowcolor{tablerow2}
512 & \textbf{6.20} & 4.16 & 72.06 \\
\ours
256 & 7.60 & 5.27 & 70.52 \\
\bottomrule
\end{tabular}
\end{small}
\end{center}
\vskip -0.1in
\end{table}

\subsection{Qualitative Analysis}
\label{sec:qualitative}

We present two examples demonstrating the ``Silent Thought, Spoken Answer'' paradigm in action. Both examples show how \method{} jointly generates internal reasoning alongside outputs, with thinking informing the final response.

\paragraph{Example 1: Speech-to-Speech (Conversational QA).}

\begin{tcolorbox}[examplebox, colback=userbox, title=User Question (Spoken)]
``Can you give me a simple way to teach my new golden retriever puppy to sit when I say the word?''
\end{tcolorbox}

\begin{tcolorbox}[examplebox, colback=thinkbox, title=Silent Thought (Text Reasoning)]
\small\itshape
The user wants a straightforward method to teach their golden retriever to sit on command. The core steps involve using treats to guide the puppy into position, introducing the verbal cue, and reinforcing the behavior through repetition. The puppy's short attention span means sessions must be brief and frequent. The response should focus on simplicity, avoiding unnecessary details like clickers or advanced techniques. The key is to align the explanation with the oral style of the reply, emphasizing clarity and ease of use.
\end{tcolorbox}

\begin{tcolorbox}[examplebox, colback=replybox, title=Spoken Answer (Audio)]
``Sure! Here's a simple way to teach your puppy to sit. First, get a treat your pup loves and stand in front of them. Hold the treat close to their nose and slowly move it up toward their head. As they follow the treat, their bottom will lower. The moment they sit, say 'sit' and give them the treat. Repeat this a few times, then take a break. Over time, say 'sit' before moving the treat, and eventually phase out the treat, using praise instead. Keep it short and fun---your pup will learn fast!''
\end{tcolorbox}

\noindent Notice how the thinking trace identifies the user's need for simplicity and structures the response accordingly. The model reasons about pedagogical approach (treat-based learning) and delivery style (oral, step-by-step), with this reasoning informing the spoken answer. This internal planning leads to a well-organized, natural-sounding response.

\paragraph{Example 2: Text-to-Text (Mathematical Reasoning).}

\begin{tcolorbox}[examplebox, colback=userbox, title=User Question (Text)]
``Solve the following math problem. Define $f(x) = 2x + 3$. Suppose that $g(x+2) = f(f(x-1) \cdot f(x+1) + f(x))$. Find $g(6)$.''
\end{tcolorbox}

\begin{tcolorbox}[examplebox, colback=thinkbox, title=Silent Thought (Text Reasoning)]
\small\itshape
1. Identify Goal: Find $g(6)$ given $f(x) = 2x+3$ and the relation for $g$. 2. Determine Input: Set $x+2 = 6$, which implies $x = 4$. 3. Substitute into Equation: Replace $x$ with 4 to get $g(6) = f(f(3) \cdot f(5) + f(4))$. 4. Evaluate Inner Functions: $f(3) = 9$, $f(5) = 13$, $f(4) = 11$. 5. Calculate Argument: $9 \cdot 13 + 11 = 128$. 6. Final Calculation: $f(128) = 259$.
\end{tcolorbox}

\begin{tcolorbox}[examplebox, colback=replybox, title=Text Answer]
``To find $g(6)$, we set $x+2 = 6$, giving $x = 4$. Substituting: $g(6) = f(f(3) \cdot f(5) + f(4))$. Using $f(x) = 2x + 3$, we compute $f(3) = 9$, $f(5) = 13$, $f(4) = 11$. Thus $g(6) = f(9 \cdot 13 + 11) = f(128) = 259$. $\boxed{259}$''
\end{tcolorbox}

\noindent Here, the thinking trace performs explicit step-by-step reasoning, breaking down the problem into manageable sub-goals. The model plans the solution strategy (identify goal, substitute, evaluate), with this reasoning informing the final answer. This structured reasoning reduces errors in multi-step mathematical problems.

\section{Conclusion}
\label{sec:conclusion}

We presented \method{}, the first diffusion-based language model for unified speech-text generation, and introduced the ``Silent Thought, Spoken Answer'' paradigm that enables speech LLMs to reason while speaking. \method{} achieves state-of-the-art speech-to-speech QA performance while maintaining strong language understanding and speech processing capabilities. The accompanying \dataset{} dataset provides the first dataset for text reasoning-augmented speech generation.

Our work demonstrates that diffusion-based generation is a viable and effective paradigm for multimodal speech-text modeling. The ability to jointly generate internal reasoning traces alongside speech output opens new possibilities for building more accurate and trustworthy spoken dialogue systems. Future work could explore the inclusion of the visual modality and extension to MoE-based models.

\section{Impact Statement}

This paper presents work whose goal is to advance the field of machine learning. There are many potential societal consequences of our work, none of which we feel must be specifically highlighted here.
\bibliography{references}

@misc{nguyen2024spiritlm,
      title={Spirit LM: Interleaved Spoken and Written Language Model}, 
      author={Tu Anh Nguyen and Benjamin Muller and Bokai Yu and Marta R. Costa-jussa and Maha Elbayad and Sravya Popuri and Christophe Ropers and Paul-Ambroise Duquenne and Robin Algayres and Ruslan Mavlyutov and Itai Gat and Mary Williamson and Gabriel Synnaeve and Juan Pino and Benoit Sagot and Emmanuel Dupoux},
      year={2024},
      eprint={2402.05755},
      archivePrefix={arXiv},
      primaryClass={cs.CL},
      url={https://arxiv.org/abs/2402.05755}, 
}

@misc{defossez2024moshi,
      title={Moshi: a speech-text foundation model for real-time dialogue}, 
      author={Alexandre Défossez and Laurent Mazaré and Manu Orsini and Amélie Royer and Patrick Pérez and Hervé Jégou and Edouard Grave and Neil Zeghidour},
      year={2024},
      eprint={2410.00037},
      archivePrefix={arXiv},
      primaryClass={eess.AS},
      url={https://arxiv.org/abs/2410.00037}, 
}

@misc{zhang2023speechgpt,
      title={SpeechGPT: Empowering Large Language Models with Intrinsic Cross-Modal Conversational Abilities}, 
      author={Dong Zhang and Shimin Li and Xin Zhang and Jun Zhan and Pengyu Wang and Yaqian Zhou and Xipeng Qiu},
      year={2023},
      eprint={2305.11000},
      archivePrefix={arXiv},
      primaryClass={cs.CL},
      url={https://arxiv.org/abs/2305.11000}, 
}

@misc{chu2024qwen2audio,
      title={Qwen2-Audio Technical Report}, 
      author={Yunfei Chu and Jin Xu and Qian Yang and Haojie Wei and Xipin Wei and Zhifang Guo and Yichong Leng and Yuanjun Lv and Jinzheng He and Junyang Lin and Chang Zhou and Jingren Zhou},
      year={2024},
      eprint={2407.10759},
      archivePrefix={arXiv},
      primaryClass={eess.AS},
      url={https://arxiv.org/abs/2407.10759}, 
}

@misc{lou2026most,
      title={MoST: Mixing Speech and Text with Modality-Aware Mixture of Experts}, 
      author={Yuxuan Lou and Kai Yang and Yang You},
      year={2026},
      eprint={2601.10272},
      archivePrefix={arXiv},
      primaryClass={cs.CL},
      url={https://arxiv.org/abs/2601.10272}, 
}

@misc{radford2023whisper,
      title={Robust Speech Recognition via Large-Scale Weak Supervision}, 
      author={Alec Radford and Jong Wook Kim and Tao Xu and Greg Brockman and Christine McLeavey and Ilya Sutskever},
      year={2022},
      eprint={2212.04356},
      archivePrefix={arXiv},
      primaryClass={eess.AS},
      url={https://arxiv.org/abs/2212.04356}, 
}

@misc{sahoo2024simple,
      title={Simple and Effective Masked Diffusion Language Models}, 
      author={Subham Sekhar Sahoo and Marianne Arriola and Yair Schiff and Aaron Gokaslan and Edgar Marroquin and Justin T Chiu and Alexander Rush and Volodymyr Kuleshov},
      year={2024},
      eprint={2406.07524},
      archivePrefix={arXiv},
      primaryClass={cs.CL},
      url={https://arxiv.org/abs/2406.07524}, 
}

@misc{shi2024simplified,
      title={Simplified and Generalized Masked Diffusion for Discrete Data}, 
      author={Jiaxin Shi and Kehang Han and Zhe Wang and Arnaud Doucet and Michalis K. Titsias},
      year={2025},
      eprint={2406.04329},
      archivePrefix={arXiv},
      primaryClass={cs.LG},
      url={https://arxiv.org/abs/2406.04329}, 
}

@misc{nie2025llada,
      title={Large Language Diffusion Models}, 
      author={Shen Nie and Fengqi Zhu and Zebin You and Xiaolu Zhang and Jingyang Ou and Jun Hu and Jun Zhou and Yankai Lin and Ji-Rong Wen and Chongxuan Li},
      year={2025},
      eprint={2502.09992},
      archivePrefix={arXiv},
      primaryClass={cs.CL},
      url={https://arxiv.org/abs/2502.09992}, 
}

@misc{dream2025,
      title={Dream 7B: Diffusion Large Language Models}, 
      author={Jiacheng Ye and Zhihui Xie and Lin Zheng and Jiahui Gao and Zirui Wu and Xin Jiang and Zhenguo Li and Lingpeng Kong},
      year={2025},
      eprint={2508.15487},
      archivePrefix={arXiv},
      primaryClass={cs.CL},
      url={https://arxiv.org/abs/2508.15487}, 
}

@misc{yang2025mmada,
      title={MMaDA: Multimodal Large Diffusion Language Models}, 
      author={Ling Yang and Ye Tian and Bowen Li and Xinchen Zhang and Ke Shen and Yunhai Tong and Mengdi Wang},
      year={2025},
      eprint={2505.15809},
      archivePrefix={arXiv},
      primaryClass={cs.CV},
      url={https://arxiv.org/abs/2505.15809}, 
}

@misc{lladav2025,
      title={LLaDA-V: Large Language Diffusion Models with Visual Instruction Tuning}, 
      author={Zebin You and Shen Nie and Xiaolu Zhang and Jun Hu and Jun Zhou and Zhiwu Lu and Ji-Rong Wen and Chongxuan Li},
      year={2025},
      eprint={2505.16933},
      archivePrefix={arXiv},
      primaryClass={cs.LG},
      url={https://arxiv.org/abs/2505.16933}, 
}

@misc{zhou2025diffa,
      title={DIFFA: Large Language Diffusion Models Can Listen and Understand}, 
      author={Jiaming Zhou and Hongjie Chen and Shiwan Zhao and Jian Kang and Jie Li and Enzhi Wang and Yujie Guo and Haoqin Sun and Hui Wang and Aobo Kong and Yong Qin and Xuelong Li},
      year={2025},
      eprint={2507.18452},
      archivePrefix={arXiv},
      primaryClass={cs.SD},
      url={https://arxiv.org/abs/2507.18452}, 
}

@misc{wei2023chain,
      title={Chain-of-Thought Prompting Elicits Reasoning in Large Language Models}, 
      author={Jason Wei and Xuezhi Wang and Dale Schuurmans and Maarten Bosma and Brian Ichter and Fei Xia and Ed Chi and Quoc Le and Denny Zhou},
      year={2023},
      eprint={2201.11903},
      archivePrefix={arXiv},
      primaryClass={cs.CL},
      url={https://arxiv.org/abs/2201.11903}, 
}

@article{deepseek2025r1,
   title={DeepSeek-R1 incentivizes reasoning in LLMs through reinforcement learning},
   volume={645},
   ISSN={1476-4687},
   url={http://dx.doi.org/10.1038/s41586-025-09422-z},
   DOI={10.1038/s41586-025-09422-z},
   number={8081},
   journal={Nature},
   publisher={Springer Science and Business Media LLC},
   author={Guo, Daya and Yang, Dejian and Zhang, Haowei and Song, Junxiao and Wang, Peiyi and Zhu, Qihao and Xu, Runxin and Zhang, Ruoyu and Ma, Shirong and Bi, Xiao and Zhang, Xiaokang and Yu, Xingkai and Wu, Yu and Wu, Z. F. and Gou, Zhibin and Shao, Zhihong and Li, Zhuoshu and Gao, Ziyi and Liu, Aixin and Xue, Bing and Wang, Bingxuan and Wu, Bochao and Feng, Bei and Lu, Chengda and Zhao, Chenggang and Deng, Chengqi and Ruan, Chong and Dai, Damai and Chen, Deli and Ji, Dongjie and Li, Erhang and Lin, Fangyun and Dai, Fucong and Luo, Fuli and Hao, Guangbo and Chen, Guanting and Li, Guowei and Zhang, H. and Xu, Hanwei and Ding, Honghui and Gao, Huazuo and Qu, Hui and Li, Hui and Guo, Jianzhong and Li, Jiashi and Chen, Jingchang and Yuan, Jingyang and Tu, Jinhao and Qiu, Junjie and Li, Junlong and Cai, J. L. and Ni, Jiaqi and Liang, Jian and Chen, Jin and Dong, Kai and Hu, Kai and You, Kaichao and Gao, Kaige and Guan, Kang and Huang, Kexin and Yu, Kuai and Wang, Lean and Zhang, Lecong and Zhao, Liang and Wang, Litong and Zhang, Liyue and Xu, Lei and Xia, Leyi and Zhang, Mingchuan and Zhang, Minghua and Tang, Minghui and Zhou, Mingxu and Li, Meng and Wang, Miaojun and Li, Mingming and Tian, Ning and Huang, Panpan and Zhang, Peng and Wang, Qiancheng and Chen, Qinyu and Du, Qiushi and Ge, Ruiqi and Zhang, Ruisong and Pan, Ruizhe and Wang, Runji and Chen, R. J. and Jin, R. L. and Chen, Ruyi and Lu, Shanghao and Zhou, Shangyan and Chen, Shanhuang and Ye, Shengfeng and Wang, Shiyu and Yu, Shuiping and Zhou, Shunfeng and Pan, Shuting and Li, S. S. and Zhou, Shuang and Wu, Shaoqing and Yun, Tao and Pei, Tian and Sun, Tianyu and Wang, T. and Zeng, Wangding and Liu, Wen and Liang, Wenfeng and Gao, Wenjun and Yu, Wenqin and Zhang, Wentao and Xiao, W. L. and An, Wei and Liu, Xiaodong and Wang, Xiaohan and Chen, Xiaokang and Nie, Xiaotao and Cheng, Xin and Liu, Xin and Xie, Xin and Liu, Xingchao and Yang, Xinyu and Li, Xinyuan and Su, Xuecheng and Lin, Xuheng and Li, X. Q. and Jin, Xiangyue and Shen, Xiaojin and Chen, Xiaosha and Sun, Xiaowen and Wang, Xiaoxiang and Song, Xinnan and Zhou, Xinyi and Wang, Xianzu and Shan, Xinxia and Li, Y. K. and Wang, Y. Q. and Wei, Y. X. and Zhang, Yang and Xu, Yanhong and Li, Yao and Zhao, Yao and Sun, Yaofeng and Wang, Yaohui and Yu, Yi and Zhang, Yichao and Shi, Yifan and Xiong, Yiliang and He, Ying and Piao, Yishi and Wang, Yisong and Tan, Yixuan and Ma, Yiyang and Liu, Yiyuan and Guo, Yongqiang and Ou, Yuan and Wang, Yuduan and Gong, Yue and Zou, Yuheng and He, Yujia and Xiong, Yunfan and Luo, Yuxiang and You, Yuxiang and Liu, Yuxuan and Zhou, Yuyang and Zhu, Y. X. and Huang, Yanping and Li, Yaohui and Zheng, Yi and Zhu, Yuchen and Ma, Yunxian and Tang, Ying and Zha, Yukun and Yan, Yuting and Ren, Z. Z. and Ren, Zehui and Sha, Zhangli and Fu, Zhe and Xu, Zhean and Xie, Zhenda and Zhang, Zhengyan and Hao, Zhewen and Ma, Zhicheng and Yan, Zhigang and Wu, Zhiyu and Gu, Zihui and Zhu, Zijia and Liu, Zijun and Li, Zilin and Xie, Ziwei and Song, Ziyang and Pan, Zizheng and Huang, Zhen and Xu, Zhipeng and Zhang, Zhongyu and Zhang, Zhen},
   year={2025},
   month=sep, pages={633–638} }

@misc{wang2026reasoninggap,
      title={Closing the Modality Reasoning Gap for Speech Large Language Models}, 
      author={Chaoren Wang and Heng Lu and Xueyao Zhang and Shujie Liu and Yan Lu and Jinyu Li and Zhizheng Wu},
      year={2026},
      eprint={2601.05543},
      archivePrefix={arXiv},
      primaryClass={cs.CL},
      url={https://arxiv.org/abs/2601.05543}, 
}

@misc{hsu2021hubert,
      title={HuBERT: Self-Supervised Speech Representation Learning by Masked Prediction of Hidden Units}, 
      author={Wei-Ning Hsu and Benjamin Bolte and Yao-Hung Hubert Tsai and Kushal Lakhotia and Ruslan Salakhutdinov and Abdelrahman Mohamed},
      year={2021},
      eprint={2106.07447},
      archivePrefix={arXiv},
      primaryClass={cs.CL},
      url={https://arxiv.org/abs/2106.07447}, 
}

@misc{jiang2025megatts3,
      title={MegaTTS 3: Sparse Alignment Enhanced Latent Diffusion Transformer for Zero-Shot Speech Synthesis}, 
      author={Ziyue Jiang and Yi Ren and Ruiqi Li and Shengpeng Ji and Boyang Zhang and Zhenhui Ye and Chen Zhang and Bai Jionghao and Xiaoda Yang and Jialong Zuo and Yu Zhang and Rui Liu and Xiang Yin and Zhou Zhao},
      year={2025},
      eprint={2502.18924},
      archivePrefix={arXiv},
      primaryClass={eess.AS},
      url={https://arxiv.org/abs/2502.18924}, 
}

@misc{microsoft2025phi4,
      title={Phi-4-Mini Technical Report: Compact yet Powerful Multimodal Language Models via Mixture-of-LoRAs}, 
      author={Microsoft and : and Abdelrahman Abouelenin and Atabak Ashfaq and Adam Atkinson and Hany Awadalla and Nguyen Bach and Jianmin Bao and Alon Benhaim and Martin Cai and Vishrav Chaudhary and Congcong Chen and Dong Chen and Dongdong Chen and Junkun Chen and Weizhu Chen and Yen-Chun Chen and Yi-ling Chen and Qi Dai and Xiyang Dai and Ruchao Fan and Mei Gao and Min Gao and Amit Garg and Abhishek Goswami and Junheng Hao and Amr Hendy and Yuxuan Hu and Xin Jin and Mahmoud Khademi and Dongwoo Kim and Young Jin Kim and Gina Lee and Jinyu Li and Yunsheng Li and Chen Liang and Xihui Lin and Zeqi Lin and Mengchen Liu and Yang Liu and Gilsinia Lopez and Chong Luo and Piyush Madan and Vadim Mazalov and Arindam Mitra and Ali Mousavi and Anh Nguyen and Jing Pan and Daniel Perez-Becker and Jacob Platin and Thomas Portet and Kai Qiu and Bo Ren and Liliang Ren and Sambuddha Roy and Ning Shang and Yelong Shen and Saksham Singhal and Subhojit Som and Xia Song and Tetyana Sych and Praneetha Vaddamanu and Shuohang Wang and Yiming Wang and Zhenghao Wang and Haibin Wu and Haoran Xu and Weijian Xu and Yifan Yang and Ziyi Yang and Donghan Yu and Ishmam Zabir and Jianwen Zhang and Li Lyna Zhang and Yunan Zhang and Xiren Zhou},
      year={2025},
      eprint={2503.01743},
      archivePrefix={arXiv},
      primaryClass={cs.CL},
      url={https://arxiv.org/abs/2503.01743}, 
}

@misc{chen2025minmo,
      title={MinMo: A Multimodal Large Language Model for Seamless Voice Interaction}, 
      author={Qian Chen and Yafeng Chen and Yanni Chen and Mengzhe Chen and Yingda Chen and Chong Deng and Zhihao Du and Ruize Gao and Changfeng Gao and Zhifu Gao and Yabin Li and Xiang Lv and Jiaqing Liu and Haoneng Luo and Bin Ma and Chongjia Ni and Xian Shi and Jialong Tang and Hui Wang and Hao Wang and Wen Wang and Yuxuan Wang and Yunlan Xu and Fan Yu and Zhijie Yan and Yexin Yang and Baosong Yang and Xian Yang and Guanrou Yang and Tianyu Zhao and Qinglin Zhang and Shiliang Zhang and Nan Zhao and Pei Zhang and Chong Zhang and Jinren Zhou},
      year={2025},
      eprint={2501.06282},
      archivePrefix={arXiv},
      primaryClass={cs.CL},
      url={https://arxiv.org/abs/2501.06282}, 
}

@misc{fang2025llamaomni2,
      title={LLaMA-Omni2: LLM-based Real-time Spoken Chatbot with Autoregressive Streaming Speech Synthesis}, 
      author={Qingkai Fang and Yan Zhou and Shoutao Guo and Shaolei Zhang and Yang Feng},
      year={2025},
      eprint={2505.02625},
      archivePrefix={arXiv},
      primaryClass={cs.CL},
      url={https://arxiv.org/abs/2505.02625}, 
}

@misc{qwen2025qwen3,
      title={Qwen3 Technical Report}, 
      author={An Yang and Anfeng Li and Baosong Yang and Beichen Zhang and Binyuan Hui and Bo Zheng and Bowen Yu and Chang Gao and Chengen Huang and Chenxu Lv and Chujie Zheng and Dayiheng Liu and Fan Zhou and Fei Huang and Feng Hu and Hao Ge and Haoran Wei and Huan Lin and Jialong Tang and Jian Yang and Jianhong Tu and Jianwei Zhang and Jianxin Yang and Jiaxi Yang and Jing Zhou and Jingren Zhou and Junyang Lin and Kai Dang and Keqin Bao and Kexin Yang and Le Yu and Lianghao Deng and Mei Li and Mingfeng Xue and Mingze Li and Pei Zhang and Peng Wang and Qin Zhu and Rui Men and Ruize Gao and Shixuan Liu and Shuang Luo and Tianhao Li and Tianyi Tang and Wenbiao Yin and Xingzhang Ren and Xinyu Wang and Xinyu Zhang and Xuancheng Ren and Yang Fan and Yang Su and Yichang Zhang and Yinger Zhang and Yu Wan and Yuqiong Liu and Zekun Wang and Zeyu Cui and Zhenru Zhang and Zhipeng Zhou and Zihan Qiu},
      year={2025},
      eprint={2505.09388},
      archivePrefix={arXiv},
      primaryClass={cs.CL},
      url={https://arxiv.org/abs/2505.09388}, 
}

@misc{xu2025qwen25omni,
      title={Qwen2.5-Omni Technical Report}, 
      author={Jin Xu and Zhifang Guo and Jinzheng He and Hangrui Hu and Ting He and Shuai Bai and Keqin Chen and Jialin Wang and Yang Fan and Kai Dang and Bin Zhang and Xiong Wang and Yunfei Chu and Junyang Lin},
      year={2025},
      eprint={2503.20215},
      archivePrefix={arXiv},
      primaryClass={cs.CL},
      url={https://arxiv.org/abs/2503.20215}, 
}

@inproceedings{panayotov2015librispeech,
  author={Panayotov, Vassil and Chen, Guoguo and Povey, Daniel and Khudanpur, Sanjeev},
  booktitle={2015 IEEE International Conference on Acoustics, Speech and Signal Processing (ICASSP)}, 
  title={Librispeech: An ASR corpus based on public domain audio books}, 
  year={2015},
  volume={},
  number={},
  pages={5206-5210},
  doi={10.1109/ICASSP.2015.7178964}}

@misc{wang2021voxpopuli,
      title={VoxPopuli: A Large-Scale Multilingual Speech Corpus for Representation Learning, Semi-Supervised Learning and Interpretation}, 
      author={Changhan Wang and Morgane Rivière and Ann Lee and Anne Wu and Chaitanya Talnikar and Daniel Haziza and Mary Williamson and Juan Pino and Emmanuel Dupoux},
      year={2021},
      eprint={2101.00390},
      archivePrefix={arXiv},
      primaryClass={cs.CL},
      url={https://arxiv.org/abs/2101.00390}, 
}

@misc{ardila2020commonvoice,
      title={Common Voice: A Massively-Multilingual Speech Corpus}, 
      author={Rosana Ardila and Megan Branson and Kelly Davis and Michael Henretty and Michael Kohler and Josh Meyer and Reuben Morais and Lindsay Saunders and Francis M. Tyers and Gregor Weber},
      year={2020},
      eprint={1912.06670},
      archivePrefix={arXiv},
      primaryClass={cs.CL},
      url={https://arxiv.org/abs/1912.06670}, 
}

@misc{hendrycks2021mmlu,
      title={Measuring Massive Multitask Language Understanding}, 
      author={Dan Hendrycks and Collin Burns and Steven Basart and Andy Zou and Mantas Mazeika and Dawn Song and Jacob Steinhardt},
      year={2021},
      eprint={2009.03300},
      archivePrefix={arXiv},
      primaryClass={cs.CY},
      url={https://arxiv.org/abs/2009.03300}, 
}

@misc{joshi2017triviaqa,
      title={TriviaQA: A Large Scale Distantly Supervised Challenge Dataset for Reading Comprehension}, 
      author={Mandar Joshi and Eunsol Choi and Daniel S. Weld and Luke Zettlemoyer},
      year={2017},
      eprint={1705.03551},
      archivePrefix={arXiv},
      primaryClass={cs.CL},
      url={https://arxiv.org/abs/1705.03551}, 
}

@misc{cobbe2021gsm8k,
      title={Training Verifiers to Solve Math Word Problems}, 
      author={Karl Cobbe and Vineet Kosaraju and Mohammad Bavarian and Mark Chen and Heewoo Jun and Lukasz Kaiser and Matthias Plappert and Jerry Tworek and Jacob Hilton and Reiichiro Nakano and Christopher Hesse and John Schulman},
      year={2021},
      eprint={2110.14168},
      archivePrefix={arXiv},
      primaryClass={cs.LG},
      url={https://arxiv.org/abs/2110.14168}, 
}

@inproceedings{berant2013webq,
    title = "Semantic Parsing on {F}reebase from Question-Answer Pairs",
    author = "Berant, Jonathan  and
      Chou, Andrew  and
      Frostig, Roy  and
      Liang, Percy",
    booktitle = "Proceedings of the 2013 Conference on Empirical Methods in Natural Language Processing",
    month = oct,
    year = "2013",
    address = "Seattle, Washington, USA",
    publisher = "Association for Computational Linguistics",
    url = "https://www.aclweb.org/anthology/D13-1160",
    pages = "1533--1544",
}

@misc{nachmani2024llamaquestions,
      title={Spoken Question Answering and Speech Continuation Using Spectrogram-Powered LLM}, 
      author={Eliya Nachmani and Alon Levkovitch and Roy Hirsch and Julian Salazar and Chulayuth Asawaroengchai and Soroosh Mariooryad and Ehud Rivlin and RJ Skerry-Ryan and Michelle Tadmor Ramanovich},
      year={2024},
      eprint={2305.15255},
      archivePrefix={arXiv},
      primaryClass={cs.CL},
      url={https://arxiv.org/abs/2305.15255}, 
}

@misc{mihaylov2018obqa,
      title={Can a Suit of Armor Conduct Electricity? A New Dataset for Open Book Question Answering}, 
      author={Todor Mihaylov and Peter Clark and Tushar Khot and Ashish Sabharwal},
      year={2018},
      eprint={1809.02789},
      archivePrefix={arXiv},
      primaryClass={cs.CL},
      url={https://arxiv.org/abs/1809.02789}, 
}

@misc{wang2025mmsu,
      title={MMSU: A Massive Multi-task Spoken Language Understanding and Reasoning Benchmark}, 
      author={Dingdong Wang and Jincenzi Wu and Junan Li and Dongchao Yang and Xueyuan Chen and Tianhua Zhang and Helen Meng},
      year={2025},
      eprint={2506.04779},
      archivePrefix={arXiv},
      primaryClass={cs.CL},
      url={https://arxiv.org/abs/2506.04779}, 
}

@misc{chen2024voicebench,
      title={VoiceBench: Benchmarking LLM-Based Voice Assistants}, 
      author={Yiming Chen and Xianghu Yue and Chen Zhang and Xiaoxue Gao and Robby T. Tan and Haizhou Li},
      year={2024},
      eprint={2410.17196},
      archivePrefix={arXiv},
      primaryClass={cs.CL},
      url={https://arxiv.org/abs/2410.17196}, 
}

@misc{allal2025smollm2,
      title={SmolLM2: When Smol Goes Big -- Data-Centric Training of a Small Language Model}, 
      author={Loubna Ben Allal and Anton Lozhkov and Elie Bakouch and Gabriel Martín Blázquez and Guilherme Penedo and Lewis Tunstall and Andrés Marafioti and Hynek Kydlíček and Agustín Piqueres Lajarín and Vaibhav Srivastav and Joshua Lochner and Caleb Fahlgren and Xuan-Son Nguyen and Clémentine Fourrier and Ben Burtenshaw and Hugo Larcher and Haojun Zhao and Cyril Zakka and Mathieu Morlon and Colin Raffel and Leandro von Werra and Thomas Wolf},
      year={2025},
      eprint={2502.02737},
      archivePrefix={arXiv},
      primaryClass={cs.CL},
      url={https://arxiv.org/abs/2502.02737}, 
}

@misc{diao2025soundmind,
      title={SoundMind: RL-Incentivized Logic Reasoning for Audio-Language Models}, 
      author={Xingjian Diao and Chunhui Zhang and Keyi Kong and Weiyi Wu and Chiyu Ma and Zhongyu Ouyang and Peijun Qing and Soroush Vosoughi and Jiang Gui},
      year={2025},
      eprint={2506.12935},
      archivePrefix={arXiv},
      primaryClass={cs.CL},
      url={https://arxiv.org/abs/2506.12935}, 
}
\bibliographystyle{icml2026}

\newpage
\appendix
\onecolumn

\section{Implementation Details}
\label{app:implementation}

This appendix provides detailed specifications referenced in the main text.

\subsection{Vocabulary Specification}
\label{app:vocab}

The unified vocabulary $\mathcal{V}$ is constructed as follows:

\begin{table}[h]
\centering
\caption{Vocabulary composition for \method{}.}
\label{tab:vocab}
\begin{tabular}{lcc}
\toprule
\textbf{Component} & \textbf{Size} & \textbf{Token ID Range} \\
\midrule
Text tokens (from LLaDA) & 126,464 & 0 -- 126,463 \\
Special tokens & 9 & 126,464 -- 126,472 \\
Speech tokens & 500 & 126,473 -- 126,972 \\
\midrule
\textbf{Total} & \textbf{126,973} & -- \\
\bottomrule
\end{tabular}
\end{table}

\paragraph{Special Tokens.}
We define 9 special tokens for task control and sequence structure:

\begin{table}[h]
\centering
\caption{Special tokens and their functions.}
\label{tab:special_tokens}
\begin{tabular}{lll}
\toprule
\textbf{Token} & \textbf{Type} & \textbf{Function} \\
\midrule
\texttt{<|t2s|>} & Task & Text-to-Speech \\
\texttt{<|asr|>} & Task & Automatic Speech Recognition \\
\texttt{<|s2s|>} & Task & Speech-to-Speech \\
\texttt{<|s2t|>} & Task & Speech-to-Text \\
\texttt{<|t2t|>} & Task & Text-to-Text \\
\texttt{<|sot|>} & Delimiter & Start of Text \\
\texttt{<|eot|>} & Delimiter & End of Text \\
\texttt{<|sos|>} & Delimiter & Start of Speech \\
\texttt{<|eos|>} & Delimiter & End of Speech \\
\bottomrule
\end{tabular}
\end{table}

\subsection{Speech Tokenization Details}
\label{app:speech_tokenization}

We use the speech tokenization pipeline from SpiritLM~\cite{nguyen2024spiritlm}:

\begin{itemize}
    \item \textbf{Encoder}: HuBERT-base model pretrained on 960 hours of LibriSpeech
    \item \textbf{Quantizer}: Linear quantizer with 500 codes trained on HuBERT features
    \item \textbf{Frame rate}: 25 Hz (one token per 40ms of audio)
    \item \textbf{Vocoder}: HiFi-GAN trained to reconstruct audio from HuBERT tokens
\end{itemize}

All components (encoder, quantizer, vocoder) are frozen during \method{} training.

\subsection{Sequence Formats}
\label{app:sequence_formats}

Complete sequence formats for all tasks, with delimiters shown explicitly:

\paragraph{Text-to-Speech (TTS):}
\begin{equation*}
[\texttt{<|t2s|>}, \texttt{<|sot|>}, \bm{t}_{\text{input}}, \texttt{<|eot|>}, \texttt{<|sos|>}, \bm{s}_{\text{output}}, \texttt{<|eos|>}]
\end{equation*}
Target: $\mathcal{Y} = \{\text{speech tokens } \bm{s}_{\text{output}}\}$

\paragraph{Automatic Speech Recognition (ASR):}
\begin{equation*}
[\texttt{<|asr|>}, \texttt{<|sos|>}, \bm{s}_{\text{input}}, \texttt{<|eos|>}, \texttt{<|sot|>}, \bm{t}_{\text{output}}, \texttt{<|eot|>}]
\end{equation*}
Target: $\mathcal{Y} = \{\text{text tokens } \bm{t}_{\text{output}}\}$

\paragraph{Speech-to-Speech with Thinking (S2S):}
\begin{equation*}
[\texttt{<|s2s|>}, \texttt{<|sos|>}, \bm{s}_{\text{user}}, \texttt{<|eos|>}, \texttt{<|sot|>}, \bm{t}_{\text{think}}, \texttt{<|eot|>}, \texttt{<|sos|>}, \bm{s}_{\text{reply}}, \texttt{<|eos|>}]
\end{equation*}
Target: $\mathcal{Y} = \{\bm{t}_{\text{think}}, \bm{s}_{\text{reply}}\}$

\paragraph{Speech-to-Text (S2T):}
\begin{equation*}
[\texttt{<|s2t|>}, \texttt{<|sos|>}, \bm{s}_{\text{user}}, \texttt{<|eos|>}, \texttt{<|sot|>}, \bm{t}_{\text{think}}, \bm{t}_{\text{reply}}, \texttt{<|eot|>}]
\end{equation*}
Target: $\mathcal{Y} = \{\bm{t}_{\text{think}}, \bm{t}_{\text{reply}}\}$

\paragraph{Text-to-Text (T2T):}
\begin{equation*}
[\texttt{<|t2t|>}, \texttt{<|sot|>}, \bm{t}_{\text{user}}, \texttt{<|eot|>}, \texttt{<|sot|>}, \bm{t}_{\text{think}}, \bm{t}_{\text{reply}}, \texttt{<|eot|>}]
\end{equation*}
Target: $\mathcal{Y} = \{\bm{t}_{\text{think}}, \bm{t}_{\text{reply}}\}$

\subsection{Training Configuration}
\label{app:training_config}

\paragraph{Stage 1: Speech-Text Alignment.}

\begin{table}[h]
\centering
\caption{Stage 1 training configuration.}
\label{tab:stage1_config}
\begin{tabular}{ll}
\toprule
\textbf{Parameter} & \textbf{Value} \\
\midrule
Base model & LLaDA-8B-Instruct \\
Tasks & TTS, ASR, LM \\
Task proportions $(p_k)$ & 0.4, 0.4, 0.2 \\
Loss coefficients $(\lambda_k)$ & 1.0, 1.0, 1.0 \\
\midrule
\multicolumn{2}{l}{\textit{Datasets}} \\
Speech (TTS/ASR) & LibriHeavy (29K hrs), VoxPopuli, Common Voice \\
Text (LM) & RefinedWeb \\
\midrule
\multicolumn{2}{l}{\textit{Optimization}} \\
Sequence length & 1024 \\
Batch size & 32 per GPU \\
GPUs & 32 $\times$ H20 \\
Training time & 22 days \\
Optimizer & AdamW \\
Learning rate & 1e-5 with cosine decay \\
Warmup steps & 1000 \\
\bottomrule
\end{tabular}
\end{table}

\paragraph{Stage 2: Instruction Following.}

\begin{table}[h]
\centering
\caption{Stage 2 training configuration.}
\label{tab:stage2_config}
\begin{tabular}{ll}
\toprule
\textbf{Parameter} & \textbf{Value} \\
\midrule
Initial checkpoint & Stage 1 model \\
Tasks & S2S, S2T, T2T, ASR, TTS \\
Task proportions $(p_k)$ & 0.3, 0.3, 0.2, 0.1, 0.1 \\
Loss coefficients $(\lambda_k)$ & 1.0 for all \\
\midrule
\multicolumn{2}{l}{\textit{Datasets}} \\
Instruction following & \dataset{} (26K samples) \\
Alignment (ASR/TTS) & Common Voice, VoxPopuli \\
\midrule
\multicolumn{2}{l}{\textit{Optimization}} \\
Sequence length & 2048 \\
GPUs & 8 \\
Optimizer & AdamW \\
Learning rate & 5e-6 \\
\bottomrule
\end{tabular}
\end{table}

\subsection{Inference Configuration}
\label{app:inference_config}

\begin{table}[h]
\centering
\caption{Inference hyperparameters.}
\label{tab:inference_config}
\begin{tabular}{ll}
\toprule
\textbf{Parameter} & \textbf{Value} \\
\midrule
Diffusion steps $T$ & 64 \\
Unmasking schedule & Linear \\
Temperature & 1.0 \\
\bottomrule
\end{tabular}
\end{table}

\section{\dataset{} Dataset Details}
\label{app:dataset}

\subsection{Source Datasets}

\dataset{} is constructed from two publicly available text QA datasets:

\begin{itemize}
    \item \textbf{Smoltalk2}~\cite{allal2025smollm2}: A conversational dataset with two subsets---SystemChat (practical knowledge and how-to questions) and EverydayConv (casual daily conversations). Licensed under Apache 2.0.
    \item \textbf{SoundMind}~\cite{diao2025soundmind}: A logical reasoning dataset focusing on multi-step reasoning tasks. Licensed under MIT.
\end{itemize}

\subsection{LLM Judge Filtering Criteria}

The quality filtering stage uses seven evaluation dimensions with the following weights:

\begin{table}[h]
\centering
\caption{LLM judge evaluation dimensions and weights.}
\begin{tabular}{llc}
\toprule
\textbf{Dimension} & \textbf{Description} & \textbf{Weight} \\
\midrule
Completeness & Response fully addresses the question & 1.0 \\
Correctness & Information is accurate and logical & 1.5 \\
Oral suitability & Appropriate for spoken delivery & 1.5 \\
Clarity & Clear and well-structured & 1.0 \\
Length appropriateness & Suitable length for speech & 0.8 \\
Visual independence & No reliance on visual elements & 1.2 \\
Conversational tone & Natural when spoken & 1.0 \\
\bottomrule
\end{tabular}
\end{table}

A sample passes if: (1) overall weighted score $\geq 3.5$, and (2) critical dimensions (correctness, oral suitability, visual independence) each score $\geq 3$.

\subsection{Audio Synthesis Configuration}

\paragraph{User Audio (MegaTTS3).}
\begin{itemize}
    \item Reference speakers: 4 voices from LibriSpeech-clean
    \item Sample rate: 24 kHz
    \item Multi-GPU parallel synthesis
\end{itemize}

\paragraph{Assistant Audio (Qwen3-Omni).}
\begin{itemize}
    \item Speaker: Single voice (``Chelsie'') for consistency
    \item Sample rate: 24 kHz
    \item Hallucination detection: WPS check (valid range: 1.5--5.5 words/sec)
    \item Minimum words for WPS check: 5 (shorter texts bypass check)
\end{itemize}

\subsection{Detailed Statistics}

\paragraph{Audio Duration Distribution.}
\begin{itemize}
    \item Minimum: 1.9 seconds
    \item Maximum: 148.5 seconds
    \item 5th percentile: 14.6 seconds
    \item 95th percentile: 83.1 seconds
\end{itemize}

\paragraph{By-Source Statistics.}

\begin{table}[h]
\centering
\caption{Per-source breakdown of \dataset{}.}
\begin{tabular}{lcccc}
\toprule
\textbf{Source} & \textbf{Count} & \textbf{Reply (words)} & \textbf{Think (words)} & \textbf{Duration (sec)} \\
\midrule
Smoltalk-SystemChat & 23,247 & 105.1 & 78.1 & 44.7 \\
Smoltalk-EverydayConv & 1,673 & 113.6 & 75.9 & 46.7 \\
SoundMind & 1,467 & 44.2 & 69.8 & 20.4 \\
\bottomrule
\end{tabular}
\end{table}

\subsection{Example Samples}

\paragraph{Example 1 (Smoltalk-SystemChat):}
\begin{quote}
\textbf{User:} ``I'm trying to set up VoiceOver on my iPhone but I'm not sure where to start. Can you help me?''

\textbf{Thinking:} ``The user needs help setting up VoiceOver on their iPhone. The main steps involve navigating to Accessibility settings, enabling VoiceOver, and learning basic gestures. The response should be structured in a clear, step-by-step manner while keeping the tone engaging and helpful...''

\textbf{Reply:} ``Sure thing! Let's get your iPhone's VoiceOver feature up and running. First, open the Settings app and scroll down to tap Accessibility. From there, select VoiceOver. Toggle the switch to On, and your iPhone will start reading the screen aloud...''
\end{quote}

\paragraph{Example 2 (SoundMind):}
\begin{quote}
\textbf{User:} ``Determine if the conclusion is entailed or not-entailed based on the premises.

Major premise: `Information poverty alleviation' refers to the government or social groups using information technology to address economic and cultural poverty caused by information blocking.

Minor premise: Which of the following is an example of information poverty alleviation?

Conclusion: A mobile phone manufacturer launched a low-cost, user-friendly smartphone for rural markets, and villagers replaced their old phones.

Is this entailed or not-entailed?''

\textbf{Thinking:} ``The answer is not entailed. The major premise specifies that information poverty alleviation must be carried out by the government or social groups, but the conclusion involves a private company. While the action might help, it doesn't meet the definition provided in the premise.''

\textbf{Reply:} ``The major premise defines information poverty alleviation as efforts by the government or social groups to use IT to solve poverty caused by information blocking. The conclusion describes a private company introducing a smartphone for rural areas. While the product may improve access to information, the actor is a commercial entity, not a government or social group. The premise emphasizes the role of the actor, which is not satisfied here. Therefore, the conclusion does not follow from the premises.''
\end{quote}

\section{Additional Experiments}
\label{app:experiments}

\subsection{Full Speech Processing Results}
\label{app:full_results}

\cref{tab:speech_processing_full} presents the complete ASR and TTS results across all four benchmarks, including CommonVoice which was omitted from the main text for space.

\begin{table}[h]
\centering
\caption{Full speech processing results (WER\%$\downarrow$) across all benchmarks. \gray{``/''} indicates unsupported.}
\label{tab:speech_processing_full}
\begin{small}
\renewcommand{\arraystretch}{1.12}
\setlength{\tabcolsep}{3pt}
\begin{tabular}{@{}l@{\hspace{4pt}}cc@{\hspace{4pt}}cc@{\hspace{4pt}}cc@{\hspace{4pt}}cc@{}}
\toprule
& \multicolumn{2}{c}{\textbf{LS-Clean}} & \multicolumn{2}{c}{\textbf{LS-Other}} & \multicolumn{2}{c}{\textbf{VoxPopuli}} & \multicolumn{2}{c}{\textbf{CommonVoice}} \\
\cmidrule(lr){2-3} \cmidrule(lr){4-5} \cmidrule(lr){6-7} \cmidrule(lr){8-9}
\textbf{Model} & ASR & TTS & ASR & TTS & ASR & TTS & ASR & TTS \\
\midrule
Qwen2-Audio & \textbf{1.8} & \gray{/} & \textbf{3.5} & \gray{/} & 7.1 & \gray{/} & 8.6 & \gray{/} \\
Phi-4-Multimodal & 2.1 & \gray{/} & 3.6 & \gray{/} & \textbf{6.3} & \gray{/} & 9.2 & \gray{/} \\
\midrule
SpeechGPT & 11.0 & 14.1 & 16.7 & 15.3 & 18.2 & 21.3 & 19.4 & 23.2 \\
SpiritLM & 6.0 & 6.7 & 11.0 & 9.5 & 14.3 & 19.4 & 15.4 & 22.4 \\
Moshi & 5.5 & 7.0 & 12.0 & 7.2 & 8.8 & 10.6 & 9.4 & 14.2 \\
MinMo & \textbf{1.8} & 6.7 & 3.9 & 7.5 & 6.7 & 10.9 & 8.0 & 13.5 \\
Llama-Omni2 & 3.5 & 10.1 & 4.0 & 9.2 & 9.5 & 12.4 & 11.3 & 17.2 \\
\midrule
\method{} & 3.0 & \textbf{6.2} & 4.3 & \textbf{6.8} & 7.1 & \textbf{10.3} & \textbf{7.6} & \textbf{11.7} \\
\bottomrule
\end{tabular}
\end{small}
\end{table}

\subsection{Speech Reasoning Results}
\label{app:mmsu_results}

\cref{tab:mmsu_full} presents results on MMSU~\cite{wang2025mmsu}, a massive multi-task spoken language understanding benchmark evaluated via VoiceBench~\cite{chen2024voicebench}. This benchmark assesses speech reasoning capabilities in the S$\to$T setting.

\begin{table}[h]
\centering
\caption{Speech reasoning results on MMSU (Accuracy\%$\uparrow$, S$\to$T only).}
\label{tab:mmsu_full}
\begin{small}
\renewcommand{\arraystretch}{1.15}
\begin{tabular}{@{}lcc@{}}
\toprule
\textbf{Model} & \textbf{Type} & \textbf{MMSU} \\
\midrule
Qwen2-Audio & AR & 35.7 \\
Phi-4-Multimodal & AR & \textbf{42.2} \\
\midrule
SpeechGPT & AR & 9.3 \\
SpiritLM & AR & 14.5 \\
Moshi & AR & 24.0 \\
MinMo & AR & \textbf{43.2} \\
Llama-Omni2 & AR & 33.4 \\
\midrule
DiFFA & Diff. & 29.6 \\
\method{} & Diff. & 39.0 \\
\bottomrule
\end{tabular}
\end{small}
\end{table}

\end{document}